\documentclass[oneside,11pt]{article} % For LaTeX2e
\renewenvironment{abstract}
  {{\centering\large\bfseries Abstract\par}\vspace{0.7ex}%
    \bgroup
       \leftskip 20pt\rightskip 20pt\small\noindent\ignorespaces}%
  {\par\egroup\vskip 0.25ex}

\newenvironment{keywords}
{\vspace{0.05in}\bgroup\leftskip 20pt\rightskip 20pt \small\noindent{\bfseries
Keywords:} \ignorespaces}%
{\par\egroup\vskip 0.25ex}

\usepackage{times}
\usepackage{graphicx}
\usepackage{algorithm,algorithmic}
\usepackage{amsfonts,amssymb,amsmath,amsthm} %sometimes replace mathtools with amsmath
\usepackage{xcolor}
\usepackage{url}
\usepackage{multirow}
\usepackage{booktabs}
\usepackage{enumitem}
\usepackage{subfig}
\usepackage[colorlinks,
            linkcolor=blue,
            citecolor=blue,
            urlcolor=magenta,
            linktocpage,
            plainpages=false]{hyperref}
\numberwithin{equation}{section}

\theoremstyle{plain}
\newtheorem{theorem}{Theorem}
\newtheorem{lemma}[theorem]{Lemma}

\theoremstyle{definition}

\DeclareMathOperator*{\Diag}{Diag}

\newcommand{\bm}{\boldsymbol{m}}

\graphicspath{{fig/}}

\title{Visual Processing by a Unified Schatten-$p$ Norm and $\ell_q$ Norm Regularized Principal Component Pursuit}
\author{
{\bf Jing Wang, Meng Wang, Xuegang Hu}\\
Department of Computer Science \\
Hefei University of Technology\\
Hefei, Anhui 230009, China
\and
{\bf Shuicheng Yan}\\
Department of Electrical and Engineering\\
National University of Singapore\\
117583, Singapore\\
}
\date{}
\begin{document}
\maketitle

\begin{abstract}
%To address the visual processing problem with corrupted data,
%Image denoising is an important field in image processing because of the typical corruption occurring in real world data.
In this paper, we propose a non-convex formulation to recover the authentic structure from the corrupted real data. Typically, the specific structure is assumed to be low rank, which holds for a wide range of data, such as images and videos. Meanwhile, the corruption is assumed to be sparse.
In the literature, such a problem is known as Robust Principal Component Analysis (RPCA), which usually recovers the low rank structure by approximating the rank function with a nuclear norm and penalizing the error by an $\ell_1$-norm.
Although RPCA is a convex formulation and can be solved effectively, the introduced norms are not tight approximations, which may cause the solution to deviate from the authentic one.
Therefore, we consider here a non-convex relaxation, consisting of a Schatten-$p$ norm and an $\ell_q$-norm that promote low rank and sparsity respectively.
We derive a proximal iteratively reweighted algorithm (PIRA) to solve the problem. Our algorithm is based on an alternating direction method of multipliers, where in each iteration we linearize the underlying objective function that allows us to have a closed form solution.
We demonstrate that solutions produced by the linearized approximation always converge and have a tighter approximation than the convex counterpart.
Experimental results on benchmarks show encouraging results of our approach.
\end{abstract}


\section{Introduction}
The popularity of webcams and mobile phone cameras has generated a large amount of visual data. However, visual data are easily corrupted by artifacts arising from imaging devices or natural factors such as illumination. The human vision system could recognize the corruption with accumulated information and knowledge. However, it will result in irrelevant or noisy information in the computer vision community. Thus, a bunch of methods have been proposed to obtain authentic data for visual processing tasks, such as image processing. Image denoising aims at reducing the noise from the original image \cite{lee1980digital} \cite{li2000spatially} \cite{zhang2013image} \cite{chou2013turbulent} \cite{shao2008heuristic}. Specifically, some approaches focus on statistical image modeling for the purpose of optimal signal representation and transmission, such as the Gaussian scale Mixture (GSM) model, the variance-adaptive model or Bayesian estimation \cite{starck2002curvelet} \cite{shao2008heuristic}. Portilla et al. presented a denoising method based on a local Gaussian scale mixture model in an overcomplete oriented pyramid representation \cite{portilla2003image}. The approaches mentioned above are based on the initial features of the visual data. Generally, better features will enhance the performance of representation. For instance, Shao et al. generated domain-adaptive global feature descriptors to obtain better performance in image classification \cite{shao2014feature}. Zhu et al. utilized weakly labeled data from other domains as the feature space for the visual categorization problem \cite{zhu2014weakly}. Based on a comprehensive feature space, some effective and promising denoising approaches are proposed by exploiting sparse and redundant representations over a trained dictionary \cite{eslami2006translation}. Elad et al. proposed the K-SVD algorithm \cite{aharon2006svd}. It was the first time that sparse modeling of image patches has been successfully applied in image denoising. Yan et al. exploited the sparsity within representation in the wavelet domain to handle high-level noises \cite{yan2013nonlocal}. One reason for the success of the algorithm is the statistical properties of noise. It is natural to assume the noise is sparse. Besides, the visual data such as images are probably of low rank structure \cite{hu2012fast}. For example, for a facial image taken under certain illumination conditions, the low-rank component captures the face, and the sparse component captures the light on the face \cite{wright2009robust}. Thus, the idea of turning the problem into a low rank matrix and a sparse matrix recovery problem has drawn considerable attention. In the following, we first describe the problem. %For a surveillance video clip, the background and the moving objects corresponding to the low-rank component and the sparse component, respectively  \cite{li2004statistical} \cite{dey2013robust} \cite{sun2013robust}.

\subsection{The Problem Description}

Suppose $X$ is an observed data matrix in $R^{m\times n}$, where $m$ is used to denote the ambient dimension of a sample and $n$ is the number of samples. The problem can be formulated as:
\begin{equation}\label{eq:opca}
\begin{split}
\min_{L,S} {\text{rank}(L)+\lambda ||S||_0}, \ \text{ s.t. } \ X = L + S,
\end{split}
\end{equation}
where $L\in R^{m\times n}$ has a low rank structure that is assumed to be the authentic structure of the observed data, and $S\in R^{m\times n}$ is assumed to be sparse representation of the noise. Rank($L$) is the rank of the matrix $L$, $||S||_0$ is the $\ell_0$-norm which counts the number of non-zero entries in $S$, and $\lambda$ is a parameter balancing the two components. The goal of the above optimization problem~\eqref{eq:opca} is called Robust Principal Component Analysis (RPCA), aiming to recover the low-rank component $L$ and sparse component $S$, with the constraint of $X=L+S$.
\subsection{The Reformulation and Solutions}
It is challenging to solve problem \eqref{eq:opca}, because rank($L$) and $||S||_0$ are both discontinuous and non-convex. In fact, it is NP-hard. A common strategy \cite{RPCA} is to relax the rank function to the convex nuclear norm  $||L||_{\ast}=\sum_{i=1}^{\min(m,n)}\sigma_i(L)$, where $\sigma_i$ denotes  the $i$-th singular value of $L$, and relax the $\ell_0$-norm to the $\ell_1$-norm $||S||_1=\sum _{ij}|S_{ij}|$, where $|S_{ij}|$ is the magnitude of the $(i, j)$-th element in $S$. Problem \eqref{eq:opca} can then be reformulated as:
\begin{equation}\label{rpca}
\begin{split}
\min_{L,S} {||L|| _{\ast}+\lambda ||S||_1}, \ \text{s.t.} \ X = L + S.
\end{split}
\end{equation}
 Cand{\`e}s et al. theoretically proved that if $L$ and $S$ satisfy certain assumptions, they can be recovered exactly via solving a convex program called Principal Component Pursuit with $\lambda=1/\sqrt{\max\{m,n\}}$ \cite{RPCA}. Unlike the formulation defined in \eqref{eq:opca}, RPCA in (\ref{rpca}) is convex, and the optimal solution is tractable. An efficient solver for (\ref{rpca}) is Alternating Direction Method (ADM) \cite{lin2010augmented} which guarantees to obtain the optimal solution. Another well-known first-order algorithm is the Accelerated Proximal Gradient (APG), which solves an unconstrained Stable Principal Component Pursuit (SPCP) problem \cite{zhou2010stable} as follows:
\begin{equation}\label{spcp}
\begin{split}
\min_{L,S} {\lambda_1||L|| _{\ast}+\lambda_2 ||S||_1+\frac{1}{2}||L+S-X||_F^2},
\end{split}
\end{equation}
where $\lambda_1>0$ and $\lambda_2>0$ are balancing parameters. APG is a fast method with a convergence rate $O(1/T^2)$, where $T$ is the number of iterations.
%As the problem \eqref{eq:opca} is NP-hard, the main strategy is to surrogate the low rank with nuclear norm and approximate the $\ell_0$-norm of sparse part with $\ell_1$-norm, known as Robust Principal Component Analysis (RPCA).
\subsection{Related Works}
As the RPCA model is capable of recovering the low rank components from grossly corrupted data and theoretical conditions to ensure the perfect recovery have been analyzed in depth, RPCA and its extensions have been applied to many applications, including background modeling \cite{RPCA},
image alignment \cite{peng2010rasl} and subspace segmentation \cite{robustlrr}. Specifically, Hui et al. presented a patch-based algorithm using low-rank matrix recovery \cite{ji2010robust}. Wang et al. studied the problem of aligning correlated images by decomposing the matrix of corrupted images as the sum of a sparse matrix of errors and a low-rank matrix of recovered aligned images \cite{wang2013robust}. Hu et al. proposed a truncated nuclear norm regularization for estimating missing values from corrupted images \cite{hu2012fast}.
%However, these methods mainly approximate the low rank rank or the sparse function separately.
%\subsection{Optimality}

There are several works aimed at improving the low-rank and sparse matrix recovery.
Mu et al. \cite{mu2011accelerated} proposed an Accelerated RPCA using random projection. Zhou and Tao \cite{zhou2011godec} developed a fast solver for low-rank and sparse matrix recovery with hard constraints on both $L$ and $S$. To alleviate the challenges raised by coherent data, most recently, Liu et al. recovered the coherent data by Low-Rank Representation (LRR) \cite{liu2013robust}. Aybat et al. developed a fast first-order algorithm to solve the SPCP problem \cite{aybat2011fast}. Fazel suggested to reformulating the rank optimization problem as a Semi-Definite Programming (SDP) problem \cite{fazel2002matrix}. An accelerated proximal gradient optimization technique was applied to solve the nuclear norm regularized least squares \cite{toh2010accelerated} \cite{ji2009accelerated}.

However, existing algorithms may lead to solutions that deviate from the original problem. Most previous works use the convex nuclear norm as a surrogate of the rank function and the $\ell_1$-norm as a surrogate of the $\ell_0$-norm, and then instead solve the new problem. But the nuclear norm is the sum of the singular values, while the rank function is the number of the non-zero singular values in which each singular value contributes equally. There are similar differences between the $\ell_0$-norm and $\ell_1$-norm when performing a theoretical analysis \cite{recht2010guaranteed}. Hence, the solution to the relaxed problem may be far from the original one.
Some researchers instead consider non-convex surrogate functions.
%\subsection{The Schatten-$p$ Norm and $\ell_q$ Norm for RPCA}
  %However, most previous works use the convex nuc These methods are all convex, to better approximate the original problem, %a wide range of approaches searching for new surrogates of the rank function and $\ell_0$-norm have been proposed.

The smoothed Schatten-$p$ norm is a popular non-convex surrogate of the rank function defined as \cite{mohan2012iterative}\cite{nie2012robust}:
\begin{equation}
\begin{split}
\ell_p(X)=~&\text{Tr}(X^TX+\epsilon I)^{p/2}\\
=~&\sum_{i=1}^{n}(\sigma_i^2(X)+\epsilon)^{p/2}
\end{split}
\end{equation}
where $I$ is the identity matrix with the same size as $X$, and $\ell_p(X)$ is differentiable for $p>0$ and nonconvex for $p<1$. %When $\epsilon = 0$, $\ell_1(X)=||X||_{\ast}$, which is also known as the $Schatten$-1 norm. % Actually, it is the $\ell_p$-norm defined on the singular values of a matrix.
Mohan and Fazel used the Schatten-$p$ norm to replace the rank function and considered the problem \cite{mohan2012iterative}: 
\begin{equation}
\begin{split}
\min~& \ell_p(X)\\
s.t.&~ A(X)=b,
\end{split}
\end{equation}
 where $A:R^{m \times n} \longrightarrow R^p$ is a linear map, and $b\in R^p$ denotes the measurements. They also proposed the Iterative Reweighted Least Squares (IRLS) algorithm for rank minimization. Under certain conditions, IRLS-1 converges to the global minimum of the smoothed nuclear norm and IRLS-$p$ converges to a stationary point of the corresponding non-convex yet smooth approximation to the rank function. Nie et al. \cite{nie2012low} proposed the extended Schatten-$p$ norm as an efficient surrogate of the rank function defined as:
 \begin{equation}
 \begin{split}
 \ell_p =& \left(  \sum_{i=1}^{\min(m,n)}\sigma_i^p\right)  ^{1/p}\\
 =&\left(  \text{Tr}(X^TX)^{p/2} \right)  ^{1/p}.
  \end{split}
 \end{equation}
They derived an efficient algorithm to solve the above problem.

For the $\ell_0$-norm, many non-convex surrogate functions have been proposed, e.g., $\ell_q$-norm with $0<q<1$ \cite{foucart2009sparsest}, and Smoothly Clipped Absolute Deviation (SCAD) \cite{fan2001variable}. Nie et al. \cite{nie2012robust} used the Alternate Direction Method (ADM) to solve a similar problem for the non-convex matrix completion problem. Cand\'es et al. \cite{candes2008enhancing} proposed an algorithm to solve the reweighted $\ell_1$ minimization problem, which could better recover the $\ell_0$-norm. The condition of sparse vector recovery has been given in \cite{foucart2009sparsest}.

The major drawback to the above approaches is that previous iteratively reweighted algorithms can only approximate either the low-rank component or the sparse one with a non-convex surrogate %squared loss or affine constraint
\cite{chenconvergence}\cite{lai2013improved}. One important reason for this is that it is difficult to solve a problem whose objective function contains two or more nonsmooth terms. Thus, in this paper, we simultaneously approximate the low rank and sparse functions with non-convex surrogates. %use the Schatten-$p$ norm as a surrogate of the rank function and the $\ell_q$-norm as a surrogate of the $\ell_0$-norm, with $0<p,q<1$.
% data, researchers can extract certain structures which are very useful in relevant domains such as image/video denoising \cite{portilla2003image} \cite{zlokolica2006wavelet}, image inpainting \cite{komodakis2006image}. For example,  % processing and video processing. However, the visual data tend to be noisy for . video data tend to be noisy for low cost cameras at high sensitivities.
%
%This problem is well known Principal Component Analysis (RPCA), aiming at obtaining an optimal low rank matrix and a sparse matrix from their sum. The RPCA has received considerable interests recently.
%  Take the problem of video denoising as an example, video denoising aiming at efficiently removing noise from all frames of a video is growing importance. A comprehensive review of the most related denoising techniques can be found in \cite{portilla2003image}. In recent years, based on the statistical properties of image noise, the idea of turning the problem of removing noise to a low rank matrix completion problem has attracted a lot of attention.
%

%Thus, we aim at solving the RPCA problem by representing the visual data as the summation of a low-rank component and a sparse component.  %Some related methods have been proposed.
\subsection{Introducing Our Approach}

In this paper, we propose a new formulation with the Schatten-$p$ norm and $\ell_q$-norm regularized Principal Component Pursuit ($p,q$-PCP) ($0<p,q<1$)
 for recovering the low-rank and sparse matrices. We also provide an algorithm to solve such a non-convex problem with two non-smooth components. %simultaneously use the Schatten-$p$ norm as a surrogate of the rank function and the $\ell_q$-norm as a surrogate of the $\ell_0$-norm, with $0<p,q<1$. %Here $p$ and $q$ are chosen separately.
 %This model is called the Schatten-$p$ norm and $\ell_q$-norm regularized Principle Component Pursuit ($p,q$-PCP) ($0 < p,q < 1$).
Empirically, our proposed Proximal Iteratively Reweighted Algorithm (PIRA) can solve $p,q$-PCP effectively without loss of efficiency. In each iteration, PIRA provides closed-form solutions which make the algorithm efficient. To the best of our knowledge, this is the first time that the $\ell_{p,q}$ norm has been used to approximate the RPCA problem. We are also the first to provide corresponding solutions. Experimental results demonstrate that the solutions can tightly approximate the RPCA problem and the objective function can converge in several iterations. The main contributions of this study are summarized as follows.
 \begin{itemize}
 \item We propose a joint Schatten-$p$ norm and $\ell_q$-norm regularized Principal Component Pursuit ($p,q$-PCP) model for low-rank and sparse matrix recovery. %$p,q$-PCP is non-convex, and it is expected that it will approximate the RPCA problem better than the convex PCP.
 \item A new Proximal Iteratively Reweighted Algorithm (PIRA) is presented to solve the $p,q$-PCP problem. We demonstrate the effectiveness and efficiency of our algorithm.%We theoretically prove that the sequence obtained by PIRA decreases the objective function value monotonically, and the limit point is a critical point.
 \item We empirically show that our solutions can approximate the original problem and the objective function will converge with a few iterations.
 \item Extensive experiments on synthetic data and real world data show that our proposed algorithm outperforms state-of-the-art algorithms.
 \end{itemize}

 \subsection{Overview of the Paper}
 The rest of the paper is organized as follows. In Section \ref{sec_2}, we give detailed information about our proposed non-convex $p,q$-PCP model and an iteratively reweighted algorithm (PIRA) to solve $p,q$-PCP. We provide a detailed analysis of the optimization algorithm in Section \ref{sec_3}. Experimental results are presented in Section \ref{sec_4}. We conclude this paper in Section \ref{sec_5}.

\section{Non-convex Principal Component Pursuit}

\label{sec_2} %涓轰簡鍜屽悗闈㈢殑supplementary material 绗﹀悎

In this section, we first present the non-convex principal component pursuit model. We then propose a new iteratively reweighted algorithm to solve the non-convex principal component pursuit problem.

\subsection{The $p,q$-PCP Model}

The motivation for approximating the rank function with $\ell_p$-norm is to obtain better empirical performance, in terms of recovering low-rank matrices, than the nuclear norm \cite{mohan2012iterative} when $0\leq p<1$. Mohan and Fazel theoretically prove that the $\ell_p$-norm is similar to the nuclear norm minimization when $p=1$ \cite{mohan2012iterative}. The $\ell_q$-norm is used with $0\leq q<1$ as a surrogate of the $\ell_0$-norm because it %approximates the $\ell_0$-norm better than the convex
generalizes and improves the $\ell_1$-norm \cite{foucart2009sparsest}. The $\ell_q$-norm degenerates into the $\ell_0$-norm when $q\rightarrow 0$. A similar property holds for the Schatten-$p$ norm as a surrogate of the rank function. It is of interest to consider the non-convex principal component pursuit by using the Schatten-$p$ norm and $\ell_q$-norm jointly:

\begin{equation}
\begin{split}\label{lpqnorm}
\min_{L,S} ~& {\lambda_1 \sum_{i=1}^{m}\sigma_i^p(L)+ \lambda_2\sum_{i=1}^m\sum_{j=1}^n|S_{ij}|^q}, \\
&\text{s.t.} \ {X=L+S,}
\end{split}
\end{equation}
where $X\in\mathbb{R}^{m\times n}$ (we assume $m\leq n$ in this paper) is the observed data matrix, and $\sigma_i(L)$, $i=1,\cdots,m$, denotes the singular values of $L$.
%$p$ is generally in the range of $[1,\infty)$. However, for the low rank constraint, we consider $0\leq p<1$ which is for measuring the rank. The value of $q$ is chosen from $(0,1)$ which is for measuring the sparsity,
$p,q\in [0,1]$, $\lambda_1, \lambda_2>0$. More generally, we further consider the stable model as follows:
%In this section, we first introduce important notations of our paper, and formulate the RPCA problem as non-convex minimization problem with no condition constraints.
%\textbf{Notification} The notations in this paper are defined as follows. Let $D\in R^{m\times n}$ be the data matrix. $D = U\Sigma V$ denotes the singular value decomposition for $D$, where $\Sigma = \text{Diag}(\sigma_i), 1\leq i\leq \min\{m,n\}$ and $\sigma_i$ is the $i$-th largest singular value of $D$. Here we will introduce some important norms will be used in this paper. The nuclear norm of $D$, $||D||_{\ast}=\sum^{\min\{m,n\}}_{i=1}\sigma_i$. The Frobenius norm of $D$ denoted as $||D||_F = \sqrt{\sum_{i,j}{D_{ij}^2}}$, the Schatten-p norm of $D$ is defined as $||D||_{S_p}^p = \sum_{i=1}^k\sigma_i^p(D)$, the $\ell_p$-norm of $D$ is $||D||_p^p=\sum_{i=1}^m\sum_{j=1}(D_{ij})^p$.
%\textbf{Our Model}
%Given a data matrix $X\in R^{m \times n}$, the goal of RPCA is to recover the low-rank matrix $L$ and a sparse matrix $S$ from $X$. We consider a non-convex formulation of PRCA with no restrictions:
%Our model is defined as:
\begin{equation}
\begin{split}\label{lpqnorm}
\min_{L,S}{\lambda_1 \sum_{i=1}^{m}\sigma_i^p(L)+ \lambda_2\sum_{i=1}^m\sum_{j=1}^n|S_{ij}|^q+\frac{1}{2}||X-L-S||_F^2}.
\end{split}
\end{equation}
When $p=q=1$, the above $p,q$-PCP model degenerates into the convex PCP as in (\ref{rpca}) or (\ref{spcp}).

It is expected that smaller values of $p$ and $q$ can help $p,q$-PCP approximate the RPCA in \eqref{eq:opca}. It is worth mentioning that many non-convex penalty functions can be applied for the non-convex principal component pursuit model. We use the Schatten-$p$ norm and the $\ell_q$-norm in this study, because compared with other non-convex penalty functions, they are matrix norms which bear more similar special properties such as the nuclear norm and $\ell_1$-norm. The low-rank matrix and the sparse matrix recovery conditions based on the Null Space Property (NSP) have been presented in previous works \cite{foucart2009sparsest}\cite{mohan2012iterative}. In fact, they are extended from the nuclear norm and the convex $\ell_1$-norm. It is easy to tune the parameters of $p$ and $q$ within $(0,1)$. Many previous works empirically showed that the $\ell_q$-norm improved the recovery performance by comparison with the convex $\ell_1$-norm \cite{candes2008enhancing}. A similar improvement was recently found in the Schatten-$p$ norm by comparison with the convex nuclear norm \cite{mohan2012iterative}. It is expected that jointly combining them in a model will surpass the recovery performance of the convex PCP.

\subsection{Proximal Iteratively Reweighted Algorithm}

In this section, we demonstrate how to solve problem (\ref{lpqnorm}) using the Schatten-$p$ norm and $\ell_q$-norm regularizers. In fact, the $\ell_q$-minimization is non-smooth, non-Lipschitz continuous, and NP-hard \cite{chen2011complexity}. We use the strategy of shifting $\sigma_i^p$ to $(\sigma_i+\varepsilon)^p$, and $|S_{ij}|^q$ to $(|S_{ij}|+\varepsilon)^q$, with $0<\varepsilon \ll 1$, and solve the relaxed problem as follows:
%The Iteratively Reweighted Least Squared (IRLS) algorithm \cite{laiimproved} smoothes the $\ell_q$-norm
%$\sum_{i,j}|S_{ij}|^q$ to $\sum_{i,j}(S_{ij}^2+\varepsilon^2)^{\frac{q}{2}}$ (with $0<\varepsilon \ll 1$), and solves the relaxed problem instead. But it may requires many iteration to obtain a sparse solution. The Iteratively Reweighted $\ell_1$ (IRL1) algorithm \cite{candes2008enhancing} instead relaxes it to $\sum_{i,j}(|S_{ij}|+\varepsilon)^q$. It requires solving a series of convex problem, and thus is not very efficient. For the Schatten-$p$ norm, the IRLS algorithm \cite{mohan2012iterative} smoothes it to $\text{Tr}(X^TX+\varepsilon I)^{\frac{1}{2}}$ (Tr($\cdot$) denotes the trace of a matrix, and $I$ is the identity matrix) and solve the relaxed problem.    We first use a common strategy which lift relax it to the following form by introducing $\varepsilon$:
\begin{equation}
\begin{split}\label{lpqnorm_r}
\min_{L,S}~~&{\lambda_1 \sum_{i=1}^{m}(\sigma_i(L)+\varepsilon)^p+ \lambda_2\sum_{i=1}^m\sum_{j=1}^n(|S_{ij}|+\varepsilon)^q}\\
&{+\frac{1}{2}||L+S-X||_F^2.}
\end{split}
\end{equation}
 $\varepsilon$ ensures that the zero singular values and nonzero entries in the sparse component have corresponding weightings. %where $w\in R^{m*1}$ and $M \in R^{m\times n}$ are the optimal weights as assigned as follows:
%To ensure that the zero singular values or the zero entries of the sparse structure have corresponding meaningful weights. we introduce $\varepsilon~(0<\varepsilon \ll 1)$.
%Thus, the problem we solve actually is:
%\begin{equation}
%\begin{split}\label{wlpq}
%\min_{L,S}~~&{\lambda_1 \sum_{i=1}^{r}w_i^k(\sigma_i(L)+\varepsilon)+ %\lambda_2\sum_{i=1}^m\sum_{j=1}^nM_{ij}^k(|S_{ij}|} \\
%&{+\varepsilon)+\frac{1}{2}||L+S-X||_F^2}\\
%\end{split}
%\end{equation}
The above problem is non-smooth and has two variables. We present a Proximal Iteratively Reweighted Algorithm (PIRA) to solve it.

Intuitively, we need to update $L$ and $S$ alternately. For fixed $S=S^k$ in the $k$-th iteration, problem (\ref{lpqnorm_r}) can be described as:
\begin{equation}
%\begin{split}
\label{org_r}
L^{k+1}=\arg\min_L {\lambda_1 \sum_{i=1}^{m}(\sigma_i(L)+\varepsilon)^p+\frac{1}{2}||L+S^k-X||_F^2.}
%\end{split}
\end{equation}
To solve the above problem, we linearize the objective function of (\ref{lpqnorm_r}) using the Taylor expansion w.r.t. $L$ at $L=L^k$ and add a proximal term. $L^{k+1}$ is then updated by minimizing the relaxed function:
\begin{equation}
\begin{split}\label{lop}
L^{k+1}=&\arg\min_L \lambda_1 \sum^m_{i=1}(\sigma_i(L^k)+\varepsilon)^p+w_i^k(\sigma_i(L)-\sigma_i(L^k))\\
&+||L^k+S^k-X||_F^2+\left\langle L^k+S^k-X, L-L^k\right\rangle \\
&+\frac{\mu_1}{2}||L-L^k||^2_F\\
=& \arg\min_L \frac{\lambda_1}{\mu_1} \sum^m_{i=1}w_i^k\sigma_i(L)\\
&+\frac{1}{2}\left\|L-\left(L^k-\frac{1}{\mu_1}(L^k+S^k-X)\right)\right\|^2_F,
\end{split}
\end{equation}
where% $w_i^k = \frac{p}{(\sigma_i(L^k)+\varepsilon)^{1-p}}, i=1,\cdots,m,$
\begin{equation}\label{eq_weightL}
w_i^k = \frac{p}{(\sigma_i(L^k)+\varepsilon)^{1-p}}, \ i=1,\cdots,m,
\end{equation}
are the weights corresponding to $L^k$. They are actually the gradients of $(\sigma_i(L)+\varepsilon)^p$ w.r.t $\sigma_i(L)$ at $L=L^k$, $i=1,\cdots,m$. The backtracking rule can be used to estimate in each iteration \cite{beck2009fast}. Note that problem (\ref{lop}) is non-convex. Fortunately, it has a closed form solution as shown in \cite{chen2012reduced}.
\begin{lemma}\label{Lem_ineq1}
%\cite[Theorem 2.3]{chen2012reduced}
Given $Y\in\mathbb{R}^{m\times n}$, $0\leq w_1\leq w_2\leq\cdots\leq w_s$ $(s=\min(m,n))$, and $\lambda>0$, the optimal solution to the following problem:
\begin{equation}
\mathcal{D}_{\lambda w}(Y)=\arg\min_Y \lambda\sum_{i=1}^sw_i\sigma_i({X})+\frac{1}{2}||{X}-{Y}||_F^2,
\end{equation}
is given by the weighted singular value threshold:
\begin{equation}
\mathcal{D}_{\lambda w}(Y)=U\mathcal{S}_{\lambda w}(\Sigma) V^T,
\end{equation}
where ${Y}={U}\Sigma{V}^T$ is the SVD of ${Y}$, and $\mathcal{S}_{\lambda {w}}(\bm{\Sigma})=\Diag\{({\Sigma}_{ii}-\lambda w_i)_+\}$.
\end{lemma}
Using Lemma \ref{Lem_ineq1}, $L^{k+1}$ can be updated by:
\begin{equation}
\begin{split}\label{wlpqL}
L^{k+1}=D_{\lambda_1w^k/\mu_1} (L^{k}-\frac{1}{\mu_1}(L^k+S^k-X)).
\end{split}
\end{equation}

\begin{algorithm}[t]
\caption{Solving Problem (\ref{lpqnorm_r}) using PIRA}
\textbf{Input:} $X\in\mathbb{R}^{m\times n}$, $0<p,q<1$, $\mu_1>1$ and $\mu_2>\frac{1}{2}$.   \\
\textbf{Initialize:} $k=0$, $w^k\in\mathbb{R}^m$, $M^k\in\mathbb{R}^{m\times n}$, $L^k\in\mathbb{R}^{m\times n}$, $S^k\in\mathbb{R}^{m\times n}$, and $\varepsilon>0$.\\
\textbf{while} not converge \textbf{do}
\begin{enumerate}
\item Update $L^{k+1}$ by %(\ref{wlpqL}).
\begin{equation*}
L^{k+1}=D_{\lambda_1 w^k/\mu_1}\left(L^k-\frac{1}{\mu_1}(L^k+S^k-X)\right).
\end{equation*}
\item Update $S^{k+1}$ by %(\ref{wlpqS}).
    \begin{equation*}
		S^{k+1}=\mathcal{S}_{\lambda_2M/\mu_2} \left(S^{k}-\frac{1}{\mu_2}(L^k+S^k-X)\right).
	\end{equation*}
\item Update the weight vector $w_i^{k+1}$, $i=1,\cdots,m$, by %(\ref{eq_weightL}).
    \begin{equation*}
		w_i^{k+1}=\frac{p}{\left(\sigma_i(L^{k+1})+\varepsilon\right)^{1-p}}.
		%w_i^{k+1}=p/(\sigma_i(L^{k+1})+\varepsilon)^{1-p},
	\end{equation*}
\item Update the weight matrix $M^{k+1}_{ij}$, $i=1,\cdots,m$, $j=1,\cdots,n$, by (\ref{eq_weightS}).
    \begin{equation*}
		M_{ij}^{k+1}=\frac{q}{(|S_{ij}^{k+1}|+\varepsilon)^{1-q}}.
	%M_{ij}^{k+1}=q/(|S_{ij}^{k+1}|+\varepsilon)^{1-q}.
	\end{equation*}
\end{enumerate}
\textbf{end while}\\
\textbf{Output}: $L^*$, $S^*$.
\label{Alg_alm}

\end{algorithm}
%\vspace{-1em}
Note that the main computation for each iteration is one SVD. The iteration is expected to obtain a better estimation of the rank successfully. Even with large rank initially, some small singular values will have large weights and will themselves become zero in the following iterations. Thus, the rank of $L$ decreases with each iteration.
\begin{figure*}
	\centering
    \includegraphics[width=0.9\textwidth]{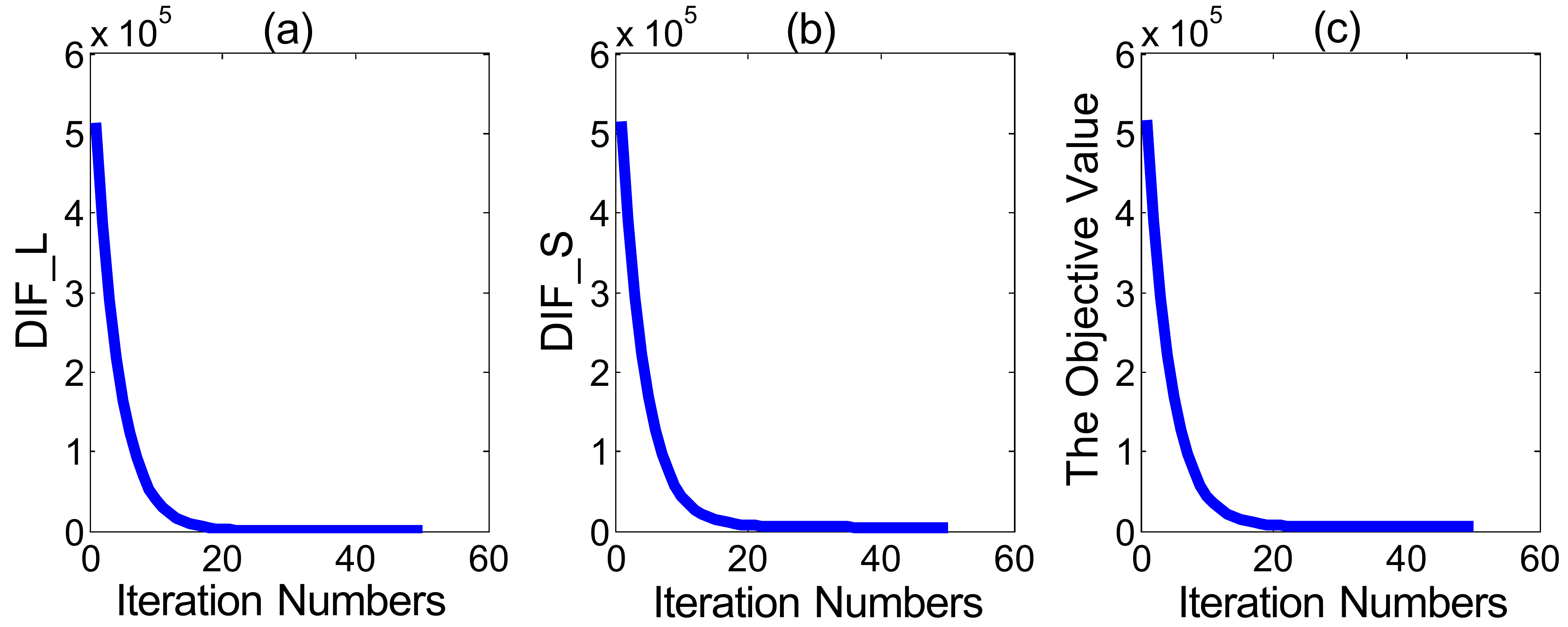}
   \caption{{The convergences of the subproblems (\ref{org_r}) %(\ref{org_r})  and problem (\ref{lop})
   and (\ref{org_s}) are shown in (a) and (b) respectively%(\ref{org_s}) and problem (\ref{updateS})
   . The convergence of the global objective function (\ref{lpqnorm}) is shown in (c).}}
	\label{conv}
	\vspace{-1em}
\end{figure*}
For fixed $L=L^k$ in the $k$-th iteration, the solution of problem (\ref{lpqnorm_r}) can be derived as:
\begin{equation}
\label{org_s}
S^{k+1}=\arg\min_S { \lambda_2\sum_{i=1}^m\sum_{j=1}^n(|S_{ij}|+\varepsilon)^q+\frac{1}{2}||L^k+S-X||_F^2.}
%\end{split}
\end{equation}
 To solve the above problem, we linearize the objective function of (\ref{lpqnorm_r}) using the Taylor expansion w.r.t $S$ at $S=S^k$ and add a proximal term. $S^{k+1}$ is then updated by minimizing the relaxed function:
\begin{equation}
\begin{split}\label{updateS}
S^{k+1}=&\arg\min_S \lambda_2\sum^m_{i=1}\sum_{j=1}^n(|S_{ij}^k|+\varepsilon)^q+M_{ij}^k(|S_{ij}|-|S_{ij}^k|)\\
&+||L^k+S^k-X||_F^2+\left\langle L^k+S^k-X, S-S^k\right\rangle \\
&+\frac{\mu_2}{2}||S-S^k||^2_F\\
=& \arg\min_S \frac{\lambda_2}{\mu_2} \sum^m_{i=1}\sum_{j=1}^nM_{ij}^k|S_{ij}|\\
&+\frac{1}{2}\left\|S-\left(S^k-\frac{1}{\mu_2}(L^k+S^k-X)\right)\right\|^2_F,
\end{split}
\end{equation}
where %$M_{ij}^k = \frac{q}{(|S_{ij}^k|+\varepsilon)^{1-q}}, i=1,\cdots,m; j=1,\cdots,n,$
\begin{equation}\label{eq_weightS}
M_{ij}^k = \frac{q}{(|S_{ij}^k|+\varepsilon)^{1-q}}, \ i=1,\cdots,m, j=1,\cdots,n,
\end{equation}
are the weights corresponding to $S^k$. They are actually the gradients of $(|S_{ij}|+\varepsilon)^q$ w.r.t. $|S_{ij}|$ at $S=S^k$, $i=1,\cdots,m$, $j=1,\cdots,n$. Note that this problem requires $O(mn)$ flops. The value of $\mu_2$ will influence the convergence of the iterations \cite{hale2008fixed}, but we empirically set it to be $2.1$ which shall hold. Note that problem (\ref{lop}) is separable. Each $S_{ij}$ can be solved separately using the following closed form solution \cite{hale2008fixed}:
%\begin{equation}
%\begin{split} \label{sop}
%S^{k+1}=& \arg\min_S \lambda_2 \sum^m_{i=1} \sum^n_{j=1} M_{ij}^k|S_{ij}|\\
%&+\left\langle L^k+S^k-X, S-S^k\right\rangle +\frac{\mu}{2}||S-S^k||^2_F\\
%=& \arg\min_L \frac{\lambda_2}{\mu} \sum^r_{i=1}w_i^k\sigma_i(L)\\
%&+\frac{1}{2}||S-(S^k-\frac{1}{\mu}(L^k+S^k-X))||^2_F
%\end{split}
%\end{equation}
%where $M_{ij} = \frac{q}{(|S_{ij}|+\varepsilon)^{1-q}}, i\in [1, \cdots, m]; j\in[1, \cdots, n]$.
%
%
%
%We introduce the theorem for solving the problem defined in Lemma \ref{Lem_ineq2}.
\begin{lemma}\label{Lem_ineq2}
Given ${y},w\in\mathbb{R}$, $w\geq0$ and $\lambda>0$, the optimal solution to the following problem
\begin{equation}
\mathcal{S}_{\lambda w}(y)=\arg\min_x \lambda w|x|+\frac{1}{2}(x-y)^2,
\end{equation}
is given by
\begin{equation}
\mathcal{S}_{\lambda w}(y)=
\begin{cases}
y-\lambda w,&\text{if}~ y>\lambda w, \\
y+\lambda w,&\text{if}~ y<-\lambda w, \\
0, & \text{otherwises}.
\end{cases}
\end{equation}
\end{lemma}
Using Lemma \ref{Lem_ineq2}, $S^{k+1}$ can be updated by
\begin{equation}\label{wlpqS}
%\begin{split}
S^{k+1}=\mathcal{S}_{\lambda_2M/\mu_2} \left(S^{k}-\frac{1}{\mu_2}(L^k+S^k-X)\right).
%\end{split}
\end{equation}

Alternately updating $L$ by (\ref{wlpqL}), $S$ by (\ref{wlpqS}) and their weights $w$ by (\ref{eq_weightL}) and $M$ by (\ref{eq_weightS}) leads to the proposed Proximal Iteratively Reweighted Algorithm (PIRA), as shown in Algorithm \ref{Alg_alm}. Note that in each iteration, PIRA solves a weighted nuclear norm minimization problem and a weighted $\ell_1$-norm minimization problem. Both have closed form solutions, and their computational costs are the same as for convex optimization. %Another advantage of PIRA is that $L$ and $S$ can be updated in parallel. Thus PIRA is an efficient method.

The detailed procedure of our algorithm is shown in Algorithm 1. We first use a common strategy which relaxes it to the form (\ref{lpqnorm_r}) by introducing $\varepsilon$, and then fix the $L$ and $S$ separately to obtain the optimal solution (step 2-3). Updated weights $w$ and $M$ are used for the next iteration based on the current solutions (step 4-7). We will provide the further analysis of our algorithm in the following section.

% The iterative scheme will converge to a local minimum of (\ref{lpqnorm})
\section{Algorithmic Analysis}
\label{sec_3}
In this section, we give a detailed analysis of our algorithm. We first illustrate that the obtained solution $L^{\ast}$ is the stationary point for problem (\ref{lop}), and the solution $S^{\ast}$ is optimal for problem (\ref{updateS}). We then show that the obtained solutions can approximate the optimal solutions of the original problems (\ref{org_r}) and (\ref{org_s}). Finally, the numerical results show that PIRA decreases the objective function (\ref{lpqnorm_r}) monotonically.

The experimental data are $X= L+S+\xi$. $L = UV^{T}$ is a rank-$10$ matrix  where $U$ and $V$ are $200\times 5$ generated by the Matlab function $\tt{randn}$. Each element of $S$ is set to zero with probability $0.8$ and non-zero entries are sampled in the interval [-5, 5] with probability $0.2$. $\xi$ is the Gaussian noise with i.i.d. $N(0, 0.01^2)$. We set $p$ to be $0.5$. As shown in the synthetic data experiment later, $p=0.5$ is representative.

%PIRA solve the problem in alternating directions with multipliers.
For the low rank part recovery, the original problem is in the form of (\ref{org_r}). With linearization, the problem (\ref{org_r}) can be derived as (\ref{lop}), which falls into the general nonconvex low-rank minimization form,
\begin{equation}
\min_{X\in R^{m\times n}} F(X) = \sum_{i=1}^{m}g_{\lambda}(\sigma_i(X)) +f(X).
\end{equation}
 We observe that the penalty function $g_{\lambda}$ in our problem (\ref{lop}) is Schatten-$p$ norm, which is continuous, concave and monotonically increasing on $[0,\infty)$. The loss function $f$ in Eqn. (\ref{lop}) is smooth and continuously differentiable with a Lipschitz continuous gradient, that is
\begin{equation}
||\nabla f(X) - \nabla f(Y)||_F \leq L(f)||X-Y||_F,
\end{equation}
where $X=L, Y = L^k-\frac{1}{\mu_1}(L^k+S^k-X)$ in the $k$-th iteration, $L(f)>0$ is the Lipschitz constant of $\nabla f$, and $\nabla$ denotes the gradient of $f$ \cite{low2014}. The solution obtained by PIRA then has the attractive properties defined in the following lemmas.

\begin{figure*}
\centering
	\subfloat[$\sigma = 0.01$]
	{
       \includegraphics[width=0.40\textwidth]{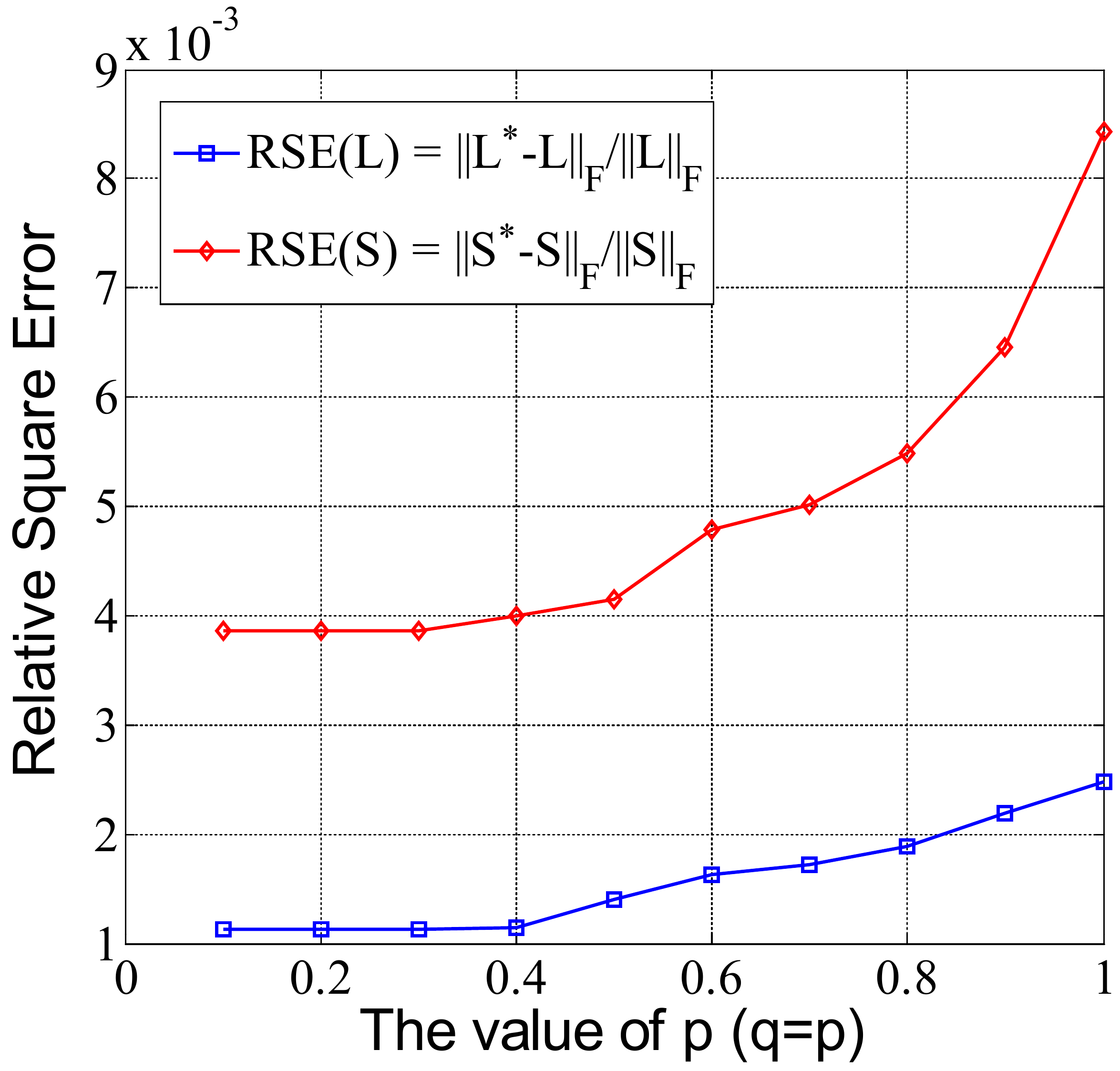}
       }
   \hfill
        \subfloat[$\sigma = 0.1$]
        {
	\includegraphics[width=0.45\textwidth]{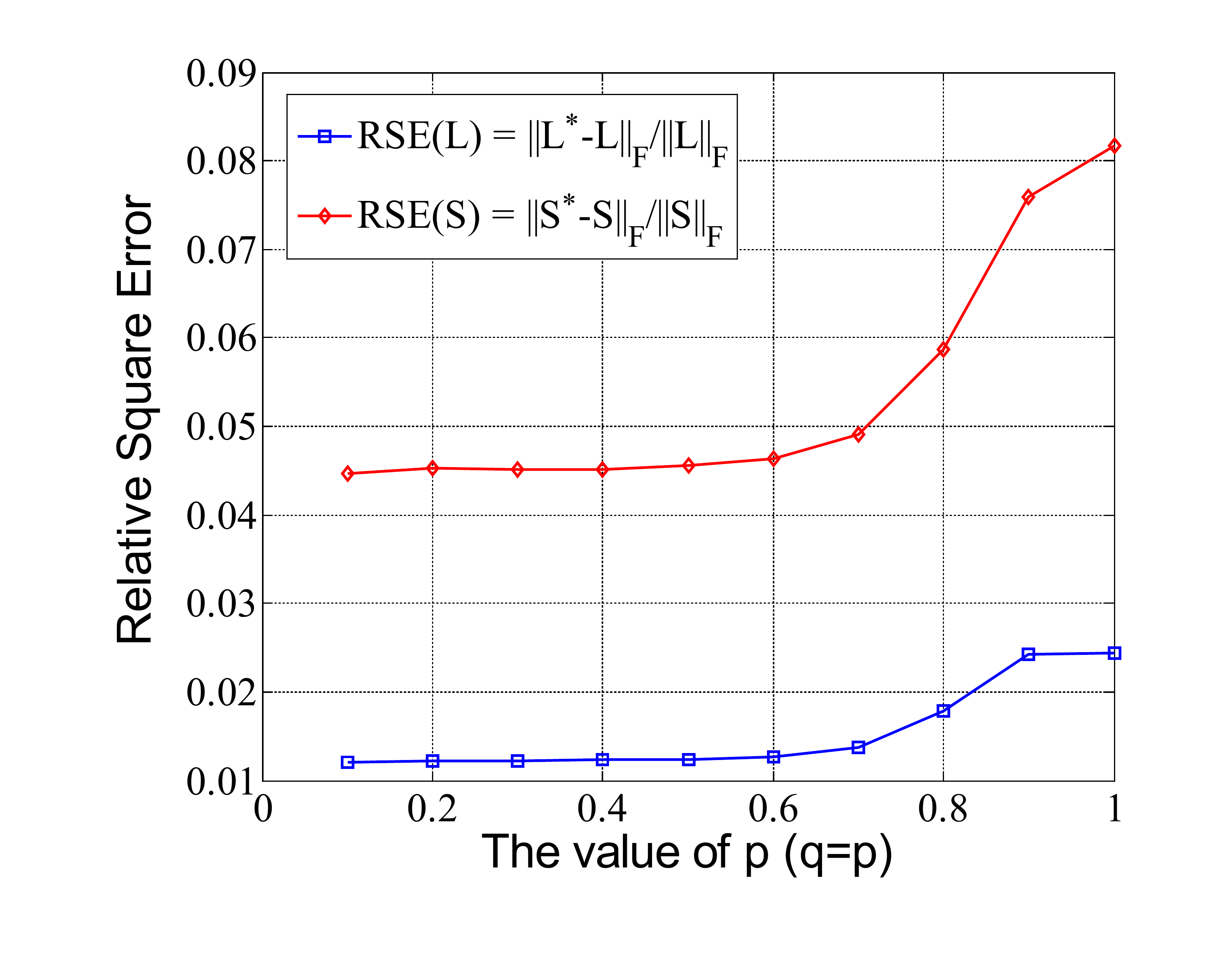}
}
%    \end{subfloat}
% \vspace{-1em}
    \caption{\small{Comparison of low-rank and sparse matrix recovery with varying noise levels $\sigma=\{0.01,0.1\}$, $n = 200$, $r=10$.}}
%    ; (a) $0.1\leqslant \lambda_1 \leqslant 0.5$, $0.001\leqslant \lambda_2 \leqslant 0.01$, (b) $5\leqslant \lambda_1 \leqslant 15$, $0.05\leqslant \lambda_2 \leqslant 0.15$
   	\label{fig_nonconfun}
	%\vspace{-1em}
\end{figure*}

 \begin{lemma}
 \label{bound}
 The sequence ${L^k}$ generated by Algorithm 1 satisfies the following properties: (1) $F(L^k)$ decreases monotonically; (2) the sequence ${L^k}$ is bounded; (3) $\sum_{k=1}^{\infty}||L^k-L^{k+1}||_F^2  \leqslant \frac{2F(L^1)}{\mu_1 -L_f}$.
\end{lemma}

\begin{lemma}
\label{converge}
Let ${L^k}$ generated by Algorithm 1 be bounded as shown in Lemma~\ref{bound}. Any accumulation point $L^{\ast}$ of ${L^k}$ is then a stationary point.
\end{lemma}

Thus, the objective function value (\ref{lop}) monotonically decreases, and any limit point of ${L^k}$ is a stationary point. Then, we empirically demonstrate the convergence of the subproblems \ref{org_r} \ref{org_s} and the global problem \ref{lpqnorm}. To this end, let $f$ denote the subproblem to recover $L$, $g$ denotes the subproblem to recover $S$, and $W$ is the global problem. In the $k$-th iteration, the objective function of $L$ problem is $f(L^{k+1};S^{k})$, the objective function of $S$ is $g(S^{k+1};L^{k})$, and the global objective function is $W(L^{k}, S^{k})$. The Figure \ref{conv} reports the convergence of the objective function values in respond to iteration numbers. Specifically, in the $k$-th iteration, the y-value of Figure \ref{conv} (a) is $||f(L^{k+1};S^{k})-f(L^{k};S^{k-1})||_F$. It is shown that the algorithm converges with limited rounds of iterations.

\section{Experiments}
\label{sec_4}
In this section, we conduct experiments on both synthetic and real visual data to validate the effectiveness of our proposed method $p,q$-PCP. In the experiment on synthetic data, we mainly discuss the influence of various $p, q$ values and noise levels. For the real-world data sets, we test our method in multiple tasks, such as image denoising and light/shadow removal.

The comparative algorithms include the classical convex solution SPCP \cite{zhou2010stable}, Non-Smooth Augmented Lagrangian algorithm (NSA) \cite{aybat2011fast} and Truncated Nuclear Norm Regularization (TNNR) \cite{hu2012fast}. We use the solver based on ADMM to solve a subproblem of TNNR in the release codes (denoted as TNNR-ADMM). As TNNR-ADMM could only recover the authentic structure, we compare our method with it in the application of image denoising. For the parameters in our algorithm, $\mu_1$ is set the same as $\mu_2$ (2.1) (empirical value). $\varepsilon$ is initialized to be $1e-3$ and decreases to $\varepsilon/p$ ($p$=1.1) after each iteration. $\lambda_1$ and $\lambda_2$ are tuned using 3-fold cross validation. Similarly, we tune the parameters for the comparative algorithms.

All the experiments are conducted with Matlab on a PC with Intel Core2 Q9550 2.83G CPU and 8G RAM.

\subsection{Synthetic Data}
%For each setting of parameters, we report the average results over 10 trails.
In this experiment, we verify the effectiveness and robustness of our algorithm by comparing with NSA and SPCP. For each setting of
parameters, we report the average result over 10 trials. We generate a rank-$r$ matrix as $L = UV^{T}$, where $U$ and $V$ are $n\times r$ generated following Gaussian distribution. The zero elements of $S$ are sampled with probability 1-$\rho_s$ and the non-zero entries are sampled in the interval [-5, 5] with probability $\rho_s$ ($\rho_s =0.2$). We further add Gaussian noise $\xi$ with i.i.d. $N(0, \sigma^2)$. $X= L+S+\xi$ is then the observed recovered matrix.

\begin{figure*}
	\subfloat[RSE ($L$)]{ %0.23
        \includegraphics[width=0.42\textwidth]{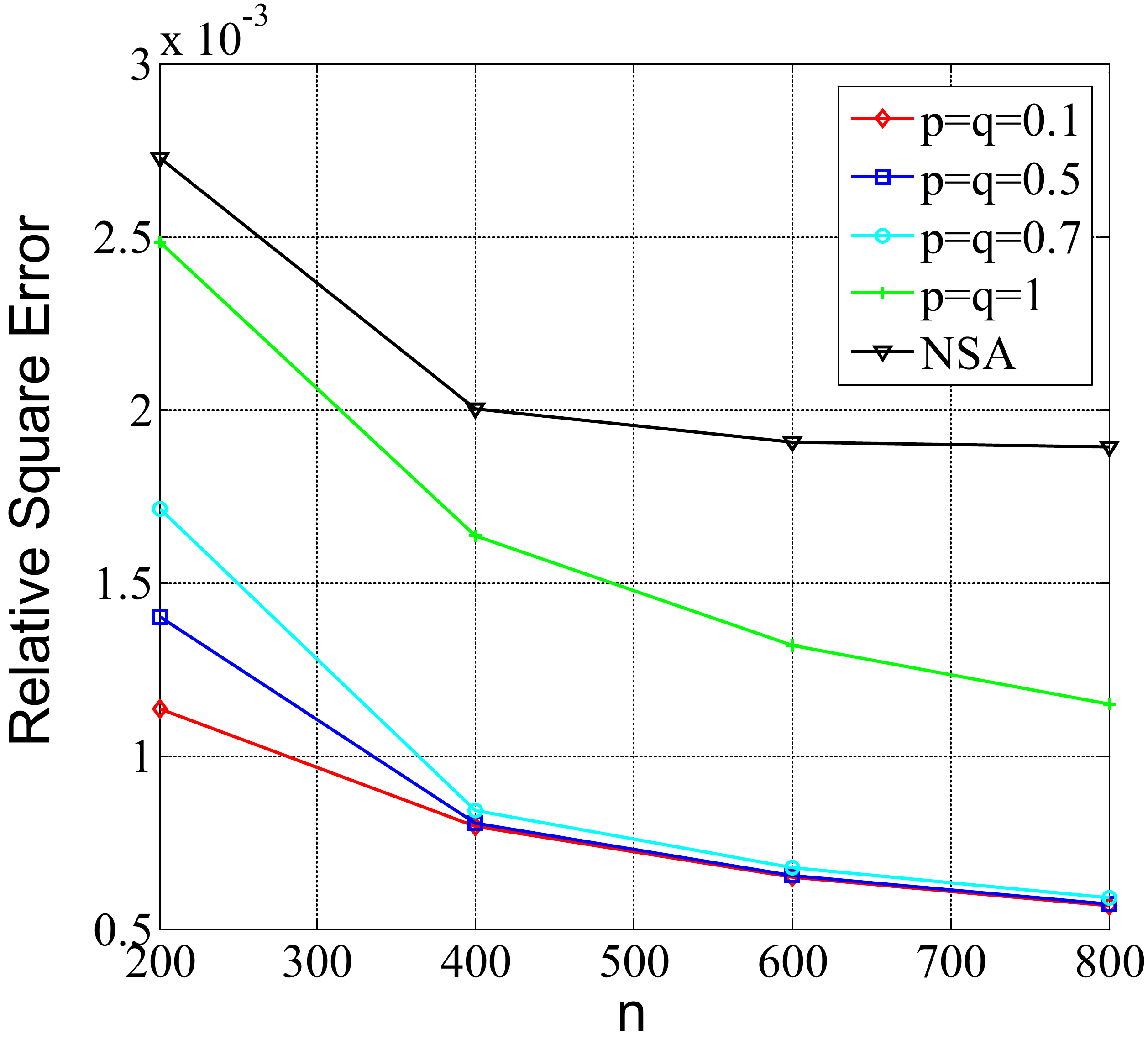}}
        \hfill
        \subfloat[RSE ($S$)]{
		\includegraphics[width=0.4\textwidth]{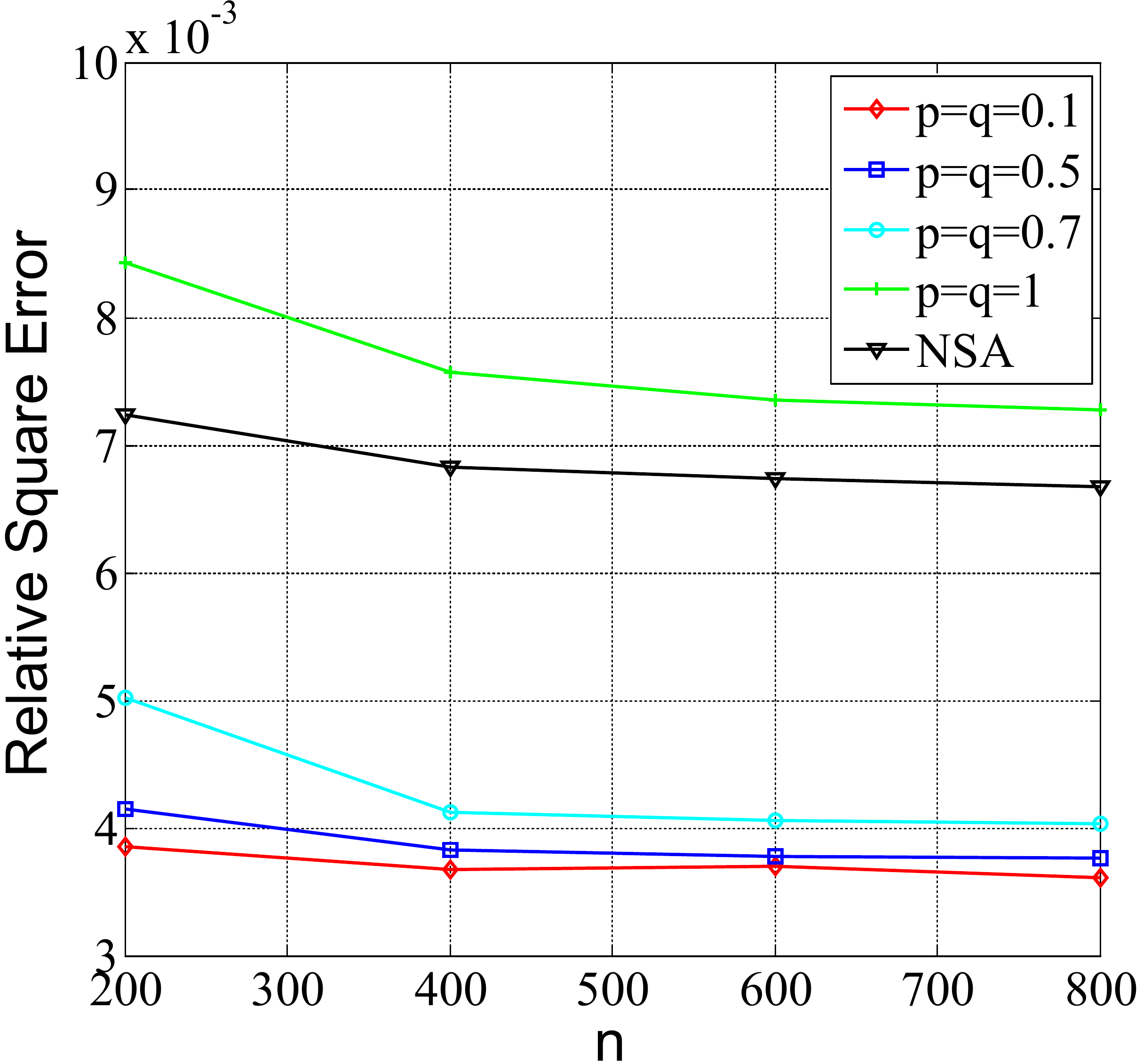}}
     %\vspace{-1em}
	    \caption{\small{Comparison of low-rank and sparse matrix recovery with varying matrix sizes $n = \{200, 400, 600, 800\}$, $\sigma=0.01$, $r=10$. In the case of $p = q = 1$, our method $p,q$-PCP is the same as SPCP.}}
	\label{fig_RS}
	%\vspace{-1em}
\end{figure*}

\begin{table*}[!t]  %(table*)
%\footnotesize
\centering
	%\vspace{-1em}	
\caption{Comparison of low-rank and sparse matrix recovery with varying underlying ranks of data.}
%\vspace{-1em}
\label{Tab_resulttab}
\centering
\begin{tabular}{|c|| c| r|r| r|r | r |}
\hline
Rank($L$)	&Algorithm	&rank($L^{\ast}$) & $ ||E||_0$	& $ \frac{||L^{\ast}-L||_F}{||L||_F}(\times 10^{-3})$ & $ \frac{||S^{\ast}-S||_F}{||S||_F}(\times 10^{-3})$  &$ \sum I(E_{ij}=E^{\ast}_{ij})$\\ \hline
%\multirow{2}*{Data Set} & \multicolumn{2}{c|}{Alpha-investing} & \multicolumn{2}{c|}{Fast-OSFS} & \multicolumn{2}{c|}{Baseline }& \multicolumn{2}{c|}{OGFS}\\
\hline

\multirow{4}*{$r$ = 5}	&NSA       & 118	   & 23041		& 3.32		& 6.50		& 0.9986 \\% \cline{2-8}
					    &SPCP      &24         &9635 	     &2.33      &7.43	     &0.8262  \\% \hline
					    &$p,q$-PCP &5		   &8023	     & 1.12 	&3.75 	 &0.9911\\ \hline
\multirow{3}*{$r$ = 15} &NSA       & 118	   &23209		 & 4.67	& 14.36				& 0.9978 \\
				    	&SPCP      &34         &10348    	&3.20    &10.87       &0.7704  \\
				    	&$p,q$-PCP &15		   &7969		&1.14		&4.06	&0.9919\\ \hline
\multirow{3}*{$r$ = 20} &NSA       &120  	   &23452    	&14.66   &50.42       &0.9941 \\
				    	&SPCP      &50		   &12569		&6.83	&23.78	 	&0.6297  \\
				    	&$p,q$-PCP &20		   &7921 	&1.25	&	4.87		&0.9870\\ \hline
\multirow{3}*{$r$ = 25} &NSA       &122    		&23825    &36.82    &142.39       &0.9775 \\
				    	&SPCP      &83			&18483	&21.95	& 84.89	&0.4249  \\
					    &$p,q$-PCP &26			& 8077	&3.25 &14.27 &0.9740\\ \hline
\end{tabular}
%\vspace{-1em}
\end{table*}

We first examine our proposed $p,q$-PCP algorithm with different $p$ and $q$. The data are generated with different noise levels, i.e., $\sigma = 0.01$ and $\sigma = 0.1$. For simplicity, we set $p=q$. When $p=q=1$, our $p,q$-PCP model is actually the SPCP model. We measure the recovery performance based on the Relative Squared Error (RSE) of the low rank part $L$ and sparse part $S$ as
\begin{equation}\label{eq_resl}
\text{RSE}({L})=\frac{||L-{L}^*||_F}{||L||_F},
\end{equation}
\begin{equation}
\text{RSE}({S})=\frac{||S-{S}^*||_F}{||S||_F},
\end{equation}
where ${L}^*$ and ${S}^*$ are the recovered matrices. The experimental results are shown in Figure \ref{fig_nonconfun}. It can be seen that $p,q$-PCP achieved better recovery performance with smaller values of $p$ and $q$. The performance of $p,q$-PCP with $p=q=0.5$ is compatible with that for $p=q=0.1$. 

To verify the effectiveness and robustness of our algorithm, we further design two experiments for comparison with other methods. One is to vary the underlying rank $r$ of the observed data, and the other to vary the dimension $n$ of the matrix. The experimental settings are as follows:
\begin{itemize}
%\item We fix $r = 10$ and $n = 200$, and vary $\sigma$ in the set $S_1$, where $S_1=\{0.01,0.1\}$;
\item We fix $\sigma = 0.01$ and $n = 200$, and vary $r$ in the set $\{5, 15, 20, 25\}$;
\item We fix $r = 10$ and $\sigma = 0.01$, and vary $n$ in the set $\{200, 400, 600, 800\}$.
\end{itemize}
Following the two directions $n$ and $r$, the experimental results for the different algorithms are shown in Table \ref{Tab_resulttab} and Figure \ref{fig_RS}. In all cases, our algorithm outperforms NSA and SPCP in terms of the rank of $L$ and the sparsity of $E$. Specifically, Figure \ref{fig_RS} records the relative square error of low-rank matrices and sparse matrices. It is obvious that our algorithm with three representative $p,q$ values all outperforms NSA and SPCP. For our own algorithm $p,q$-PCP, the differences between $p,q$ = 0.1 and $p,q$ = 0.5 are limited. Thus, in the following experiments, we adopt $p,q = 0.5$ to test our algorithm. Table \ref{Tab_resulttab} shows comprehensive results of the recovery errors related to $L$ and $E$ (column 4-5), and the accuracy of the captured sparse location (column 6). $L^{\ast}$ and $S^{\ast}$ are the solutions obtained using different algorithms, and $L$, $S$ are the groundtruth matrices. $||E^{\ast}||_0$ represents the number of non-zero entries of $E^*$. $\sum I(E_{ij}=E^{\ast}_{ij})$ records the percentage of correctly located entries. From the results shown in Table \ref{Tab_resulttab}, we see that our algorithm and NSA both perform much better than SPCP in capturing the sparse locations in matrix $E$. For recovering the low-rank and sparse matrix, our algorithm obtains the best approximation. In particular, in the case of $r$ = 25, compared with NSA ($36.82\times 10^{-3}$) and SPCP ($21.95\times 10^{-3}$), we obtain $3.25\times 10^{-3}$ relative error of $L$. Furthermore, our algorithm achieves $14.27\times 10^{-3}$ RSE of $E$, which is much better than NSA ($142.39\times 10^{-3}$) and SPCP ($84.89\times 10^{-3}$).

\begin{figure*}
	\centering
	\subfloat[Original]{
	        \includegraphics[width=0.1615\textwidth]{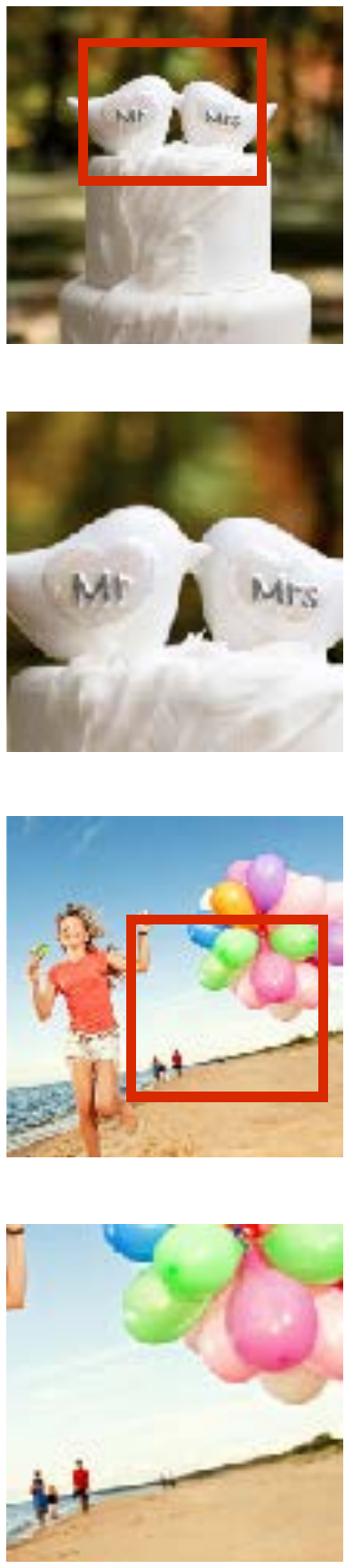}}
	\subfloat[Noise]{
	        \includegraphics[width=0.1626\textwidth]{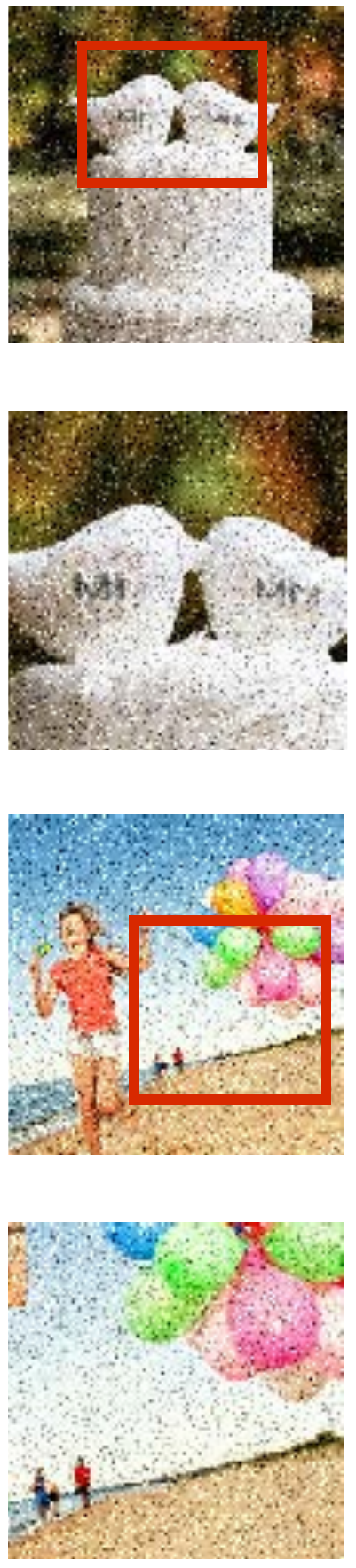}}
    \subfloat[TNNR-ADMM]{
	        \includegraphics[width=0.1616\textwidth]{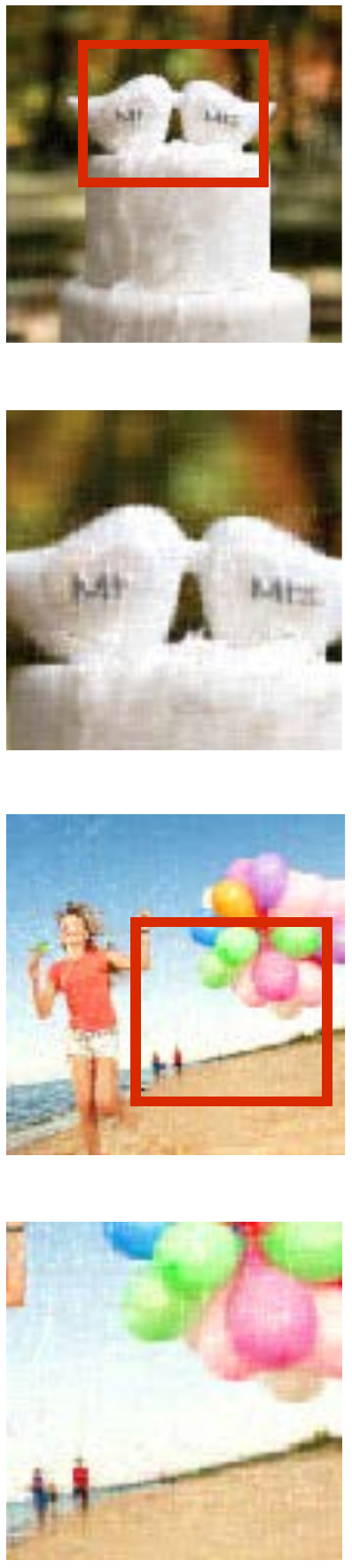}}
	\subfloat[NSA]{
	        \includegraphics[width=0.16\textwidth]{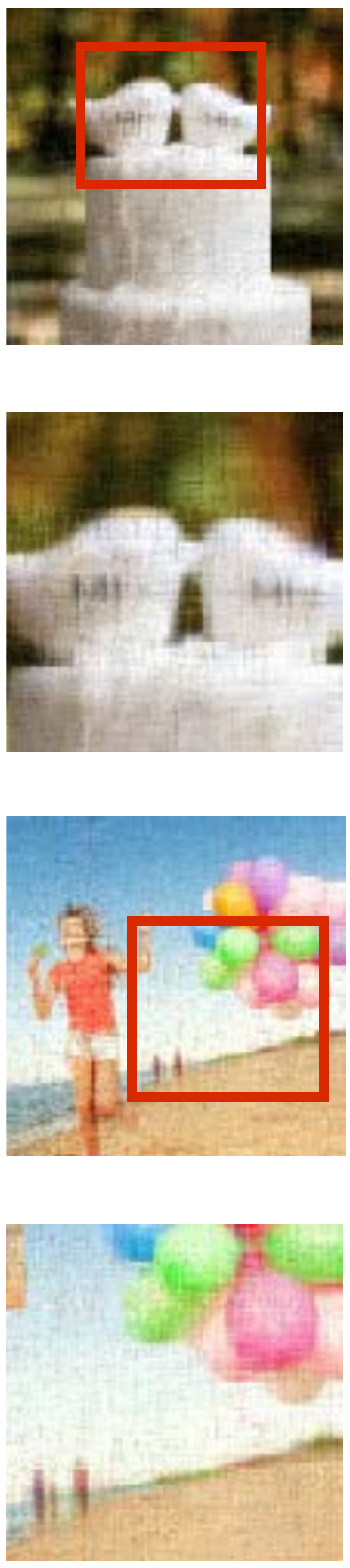}}
	\subfloat[SPCP]{
	        \includegraphics[width=0.1605\textwidth]{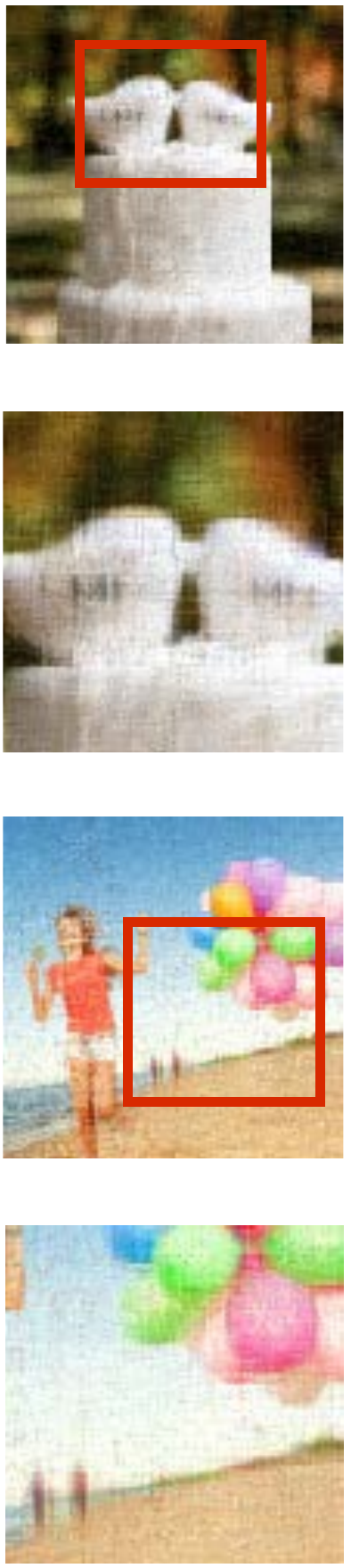}}
	\subfloat[$p,q$-PCP]{
	        \includegraphics[width=0.159\textwidth]{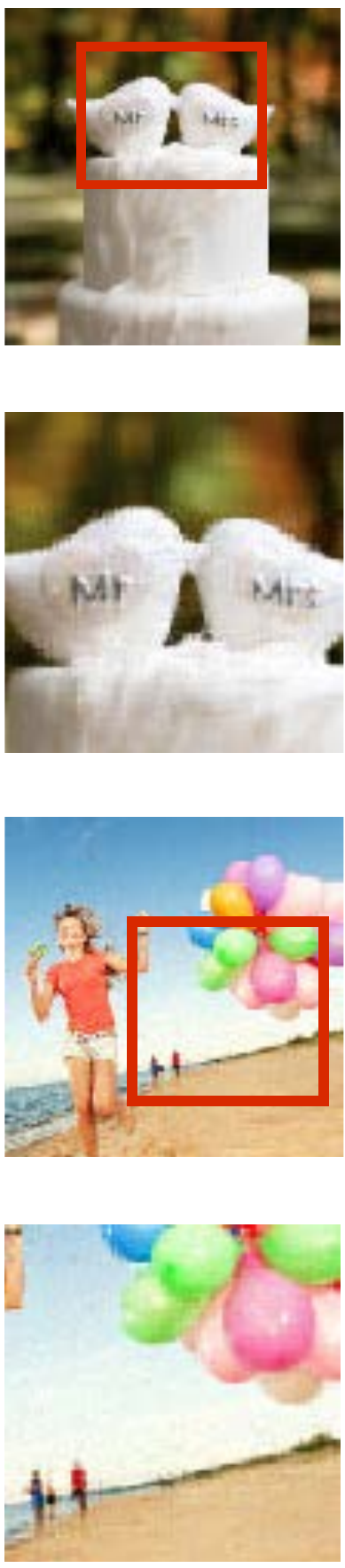}}
    \caption{Comparison of image recovery using different low rank approximation algorithms. The images in the second and last row are the amplified patches circled by the red
    Q4 line in the previous row. }
    \label{fig_pic_pic}
    \vspace{-1.5em}
\end{figure*}
\vspace{-1em}

\begin{figure*} %image_gauss_rms.pdf
	\subfloat[]{
        \includegraphics[width=0.46\textwidth]{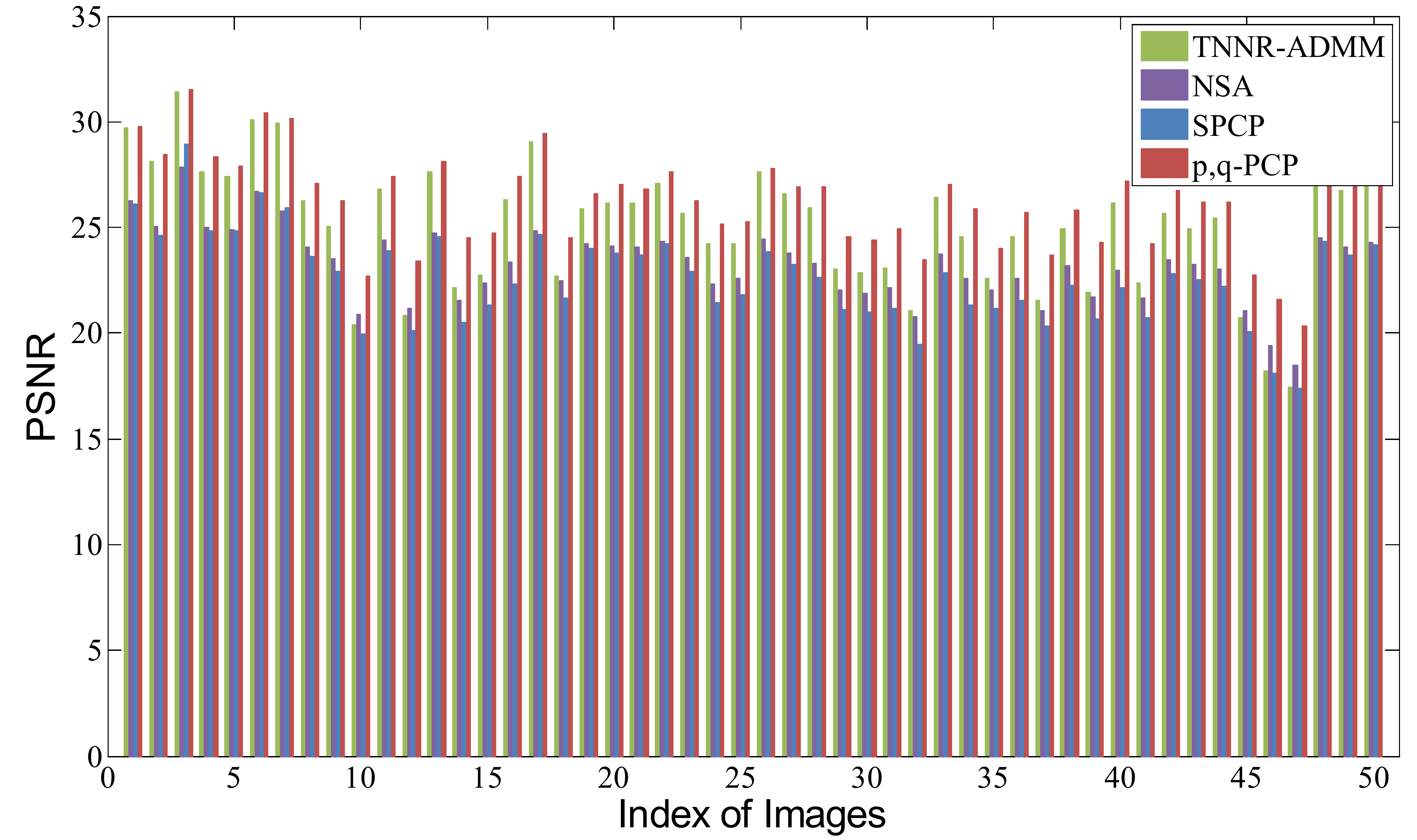}}
	\subfloat[]{ %image_gauss_psnr
        \includegraphics[width=0.49\textwidth]{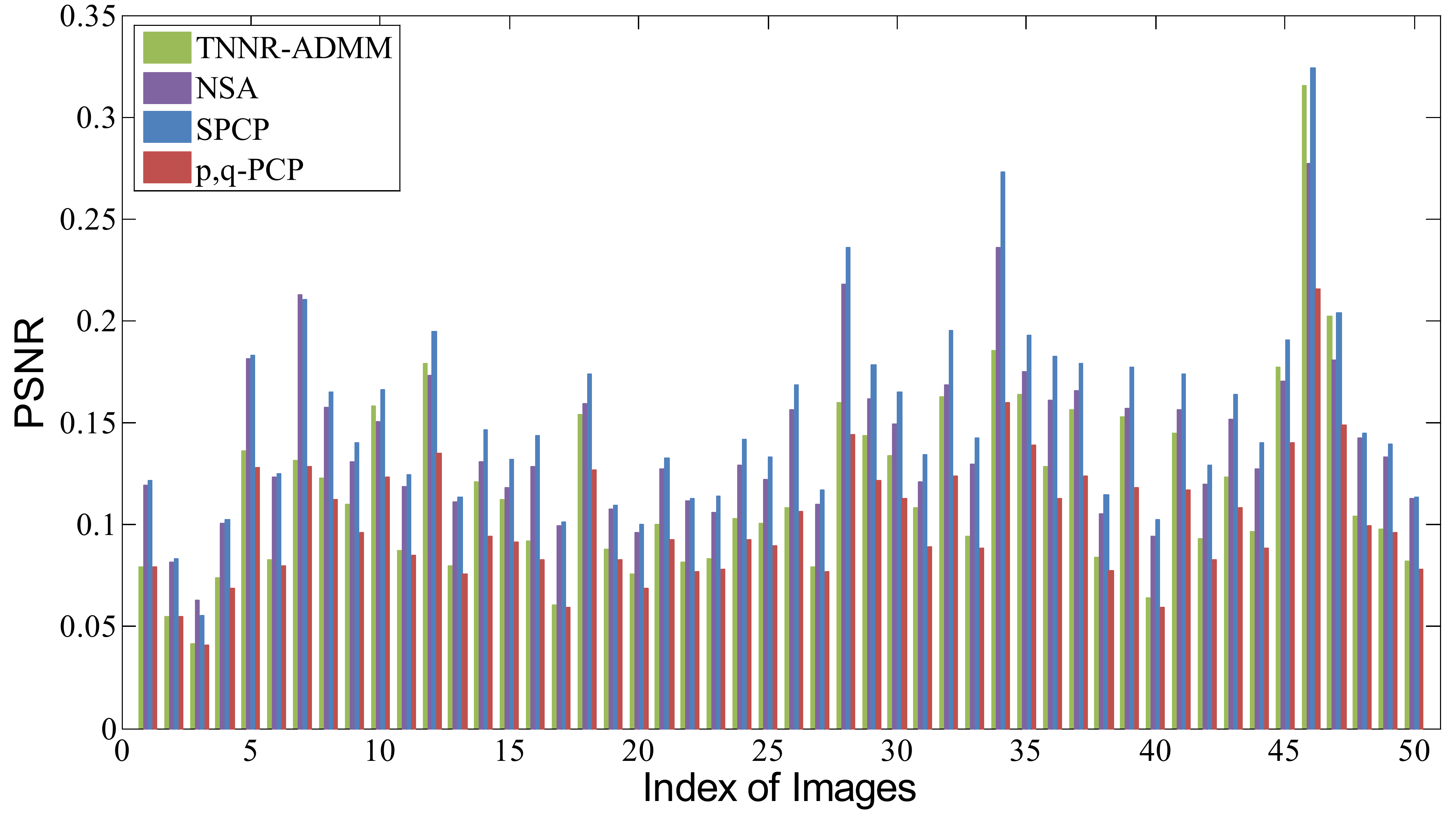}}
    % \vspace{-1em}
	    \caption{\small{Comparison of the relative square error (a) and PSNR values (b) on the 50 natural images.}}
	     \label{fig_pic_psnr_rms}
	  \vspace{-1em}
\end{figure*}

\subsection{Image Denoising}
The real images may be corrupted by different types of noise. In this experiment, we apply the low-rank approximation algorithms for image denoising. We download 50 images from the Google image search engine. The images are with three channels and of the same size $300\times 300$. We add Gaussian noise $N(0,0.2^2)$ to $50\%$ of the pixels of each image. Note that the color image consists of three channels: $L_{\text{red}}$, $L_{\text{green}}$ and $L_{\text{blue}}\in R^{m\times n}$. We implement the principal component analysis algorithms for each channel respectively. The image is then recovered by combining the recovered results from the three channels. Some recovered images are shown in Figure \ref{fig_pic_pic}. It can be seen that our method achieves the best image recovery performance. The images recovered by NSA and SPCP are blurred and some important details are missing. The recovered images of TNNR-ADMM are much clear than NSA and SPCP, but still not as good as our method.

We measure the recovery performance based on the RSE(L) defined in (\ref{eq_resl}) and the PSNR (Peak Signal Noise Ratio) \cite{huynh2008scope} value. Figure~\ref{fig_pic_psnr_rms} plots the RSE and PSNR values on 50 tested images. It can be seen that our algorithm obtains the highest PSNR values and the smallest RSE for all the images. Such results indicate that the our low-rank approximation is better than the traditional nuclear norm heuristic in this situation. Though TNNR-ADMM is non-convex method, it is still inferior to our method. It demonstrates the effectiveness of our model and the optimization method.

\begin{figure*}
   \subfloat[Original]{
        \includegraphics[width=0.14\textwidth]{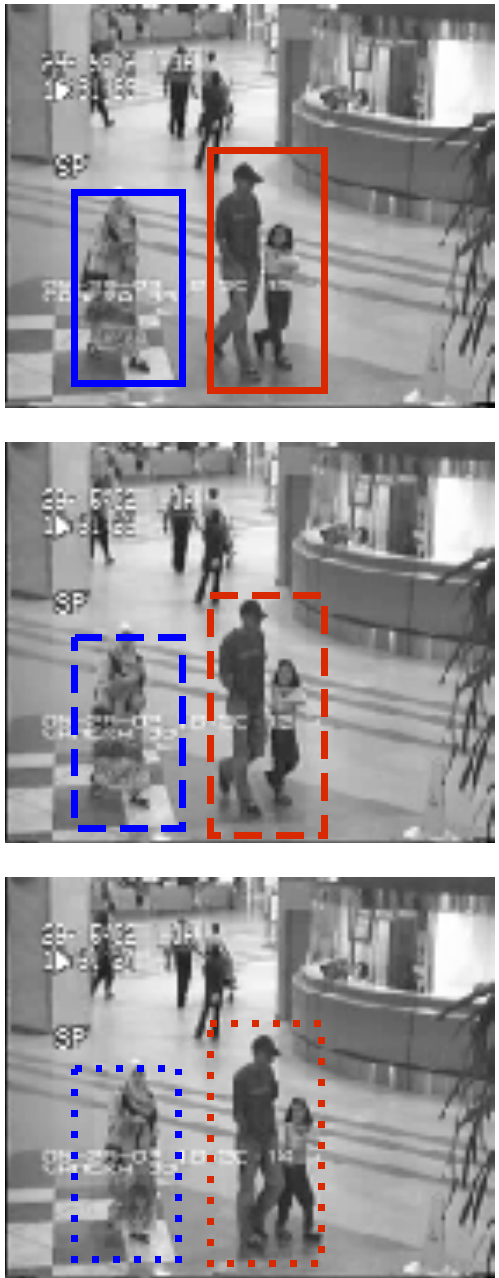}}
     \subfloat[NSA]{
        \includegraphics[width=0.30\textwidth]{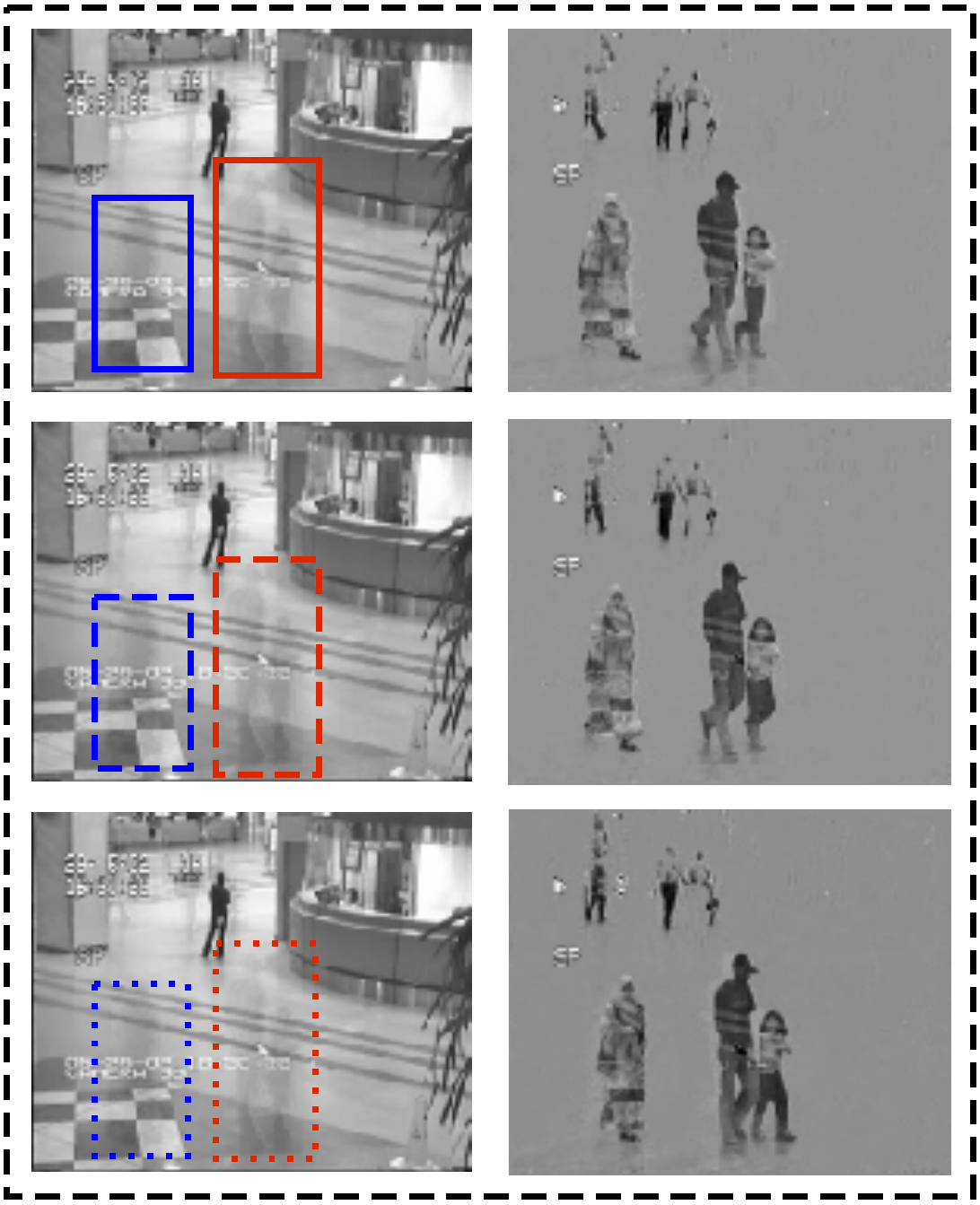}}
    \subfloat[SPCP]{
		\includegraphics[width=0.30\textwidth]{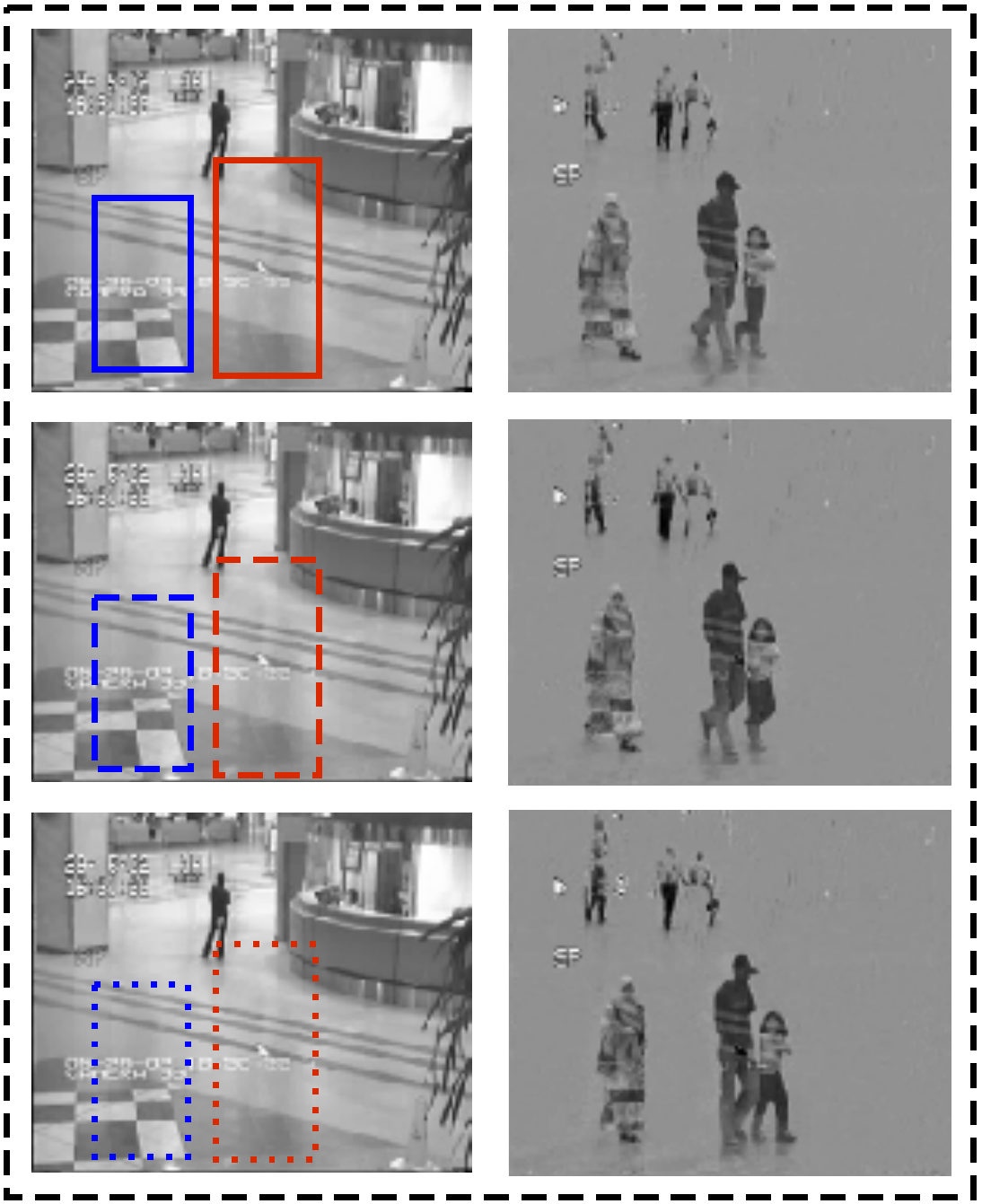}}
	\subfloat[$p,q$-PCP]{
        \includegraphics[width=0.30\textwidth]{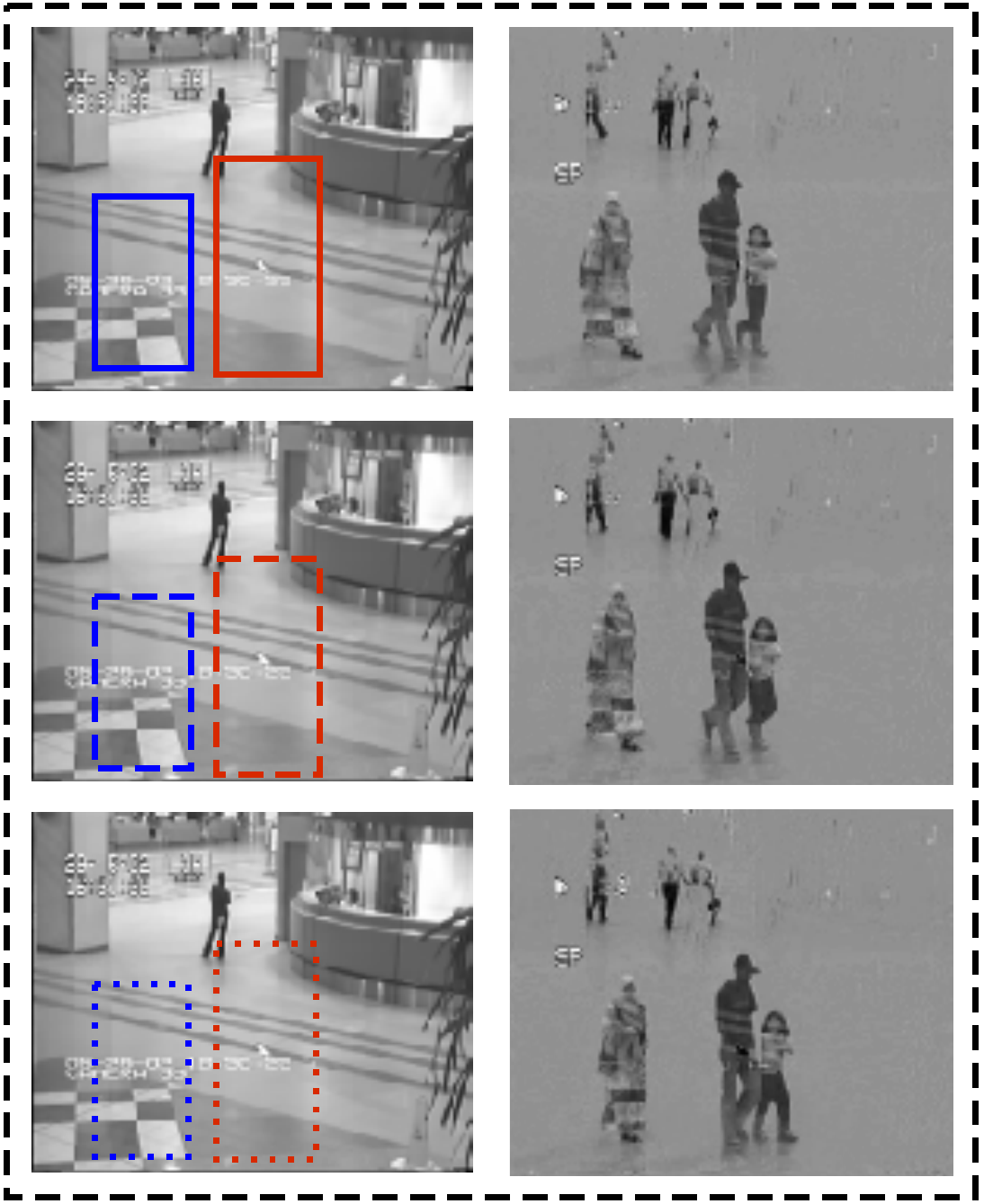}}
    \caption{Background extraction using different algorithms. Because of limited space, we only plot the results of three frames (corresponding to three rows).}
	\label{fig_video}
	\vspace{-1em}
\end{figure*}

\subsection{Other Applications}

To test the generalization of our method, we design two algorithms for two more applications, separation of foreground and background for a surveillance video and removal of light/shadow from facial images.

Specifically, we first apply different approaches to background separation and detection of objects in the foreground in an airport surveillance video \cite{li2004statistical}. It contains a sequence of 201 grayscale frames of 144$\times$176 pixels during an observation time period. The size of the observed data is $X\in R^{25344\times 201}$.

Figure \ref{fig_video} shows the background extraction results using NSA, SPCP and $p,q$-PCP algorithms. It can be seen that all three methods are able to separate the background and foreground. To further compare the background recovery results clearly, we mark the ghosting parts of the NSA with rectangles, and mark the same parts with the same rectangles in the original frame, as well as in the recovered background of SPCP and $p,q$-PCP. Although the SPCP has a better recovered background than NSA, there are ghosting shadows in the recovered background. Our algorithm best recovers the background, removing almost all the shadows. 

\begin{figure*}
\centering
   \subfloat[Subject 1]{%0.23
        \includegraphics[width=0.35\textwidth]{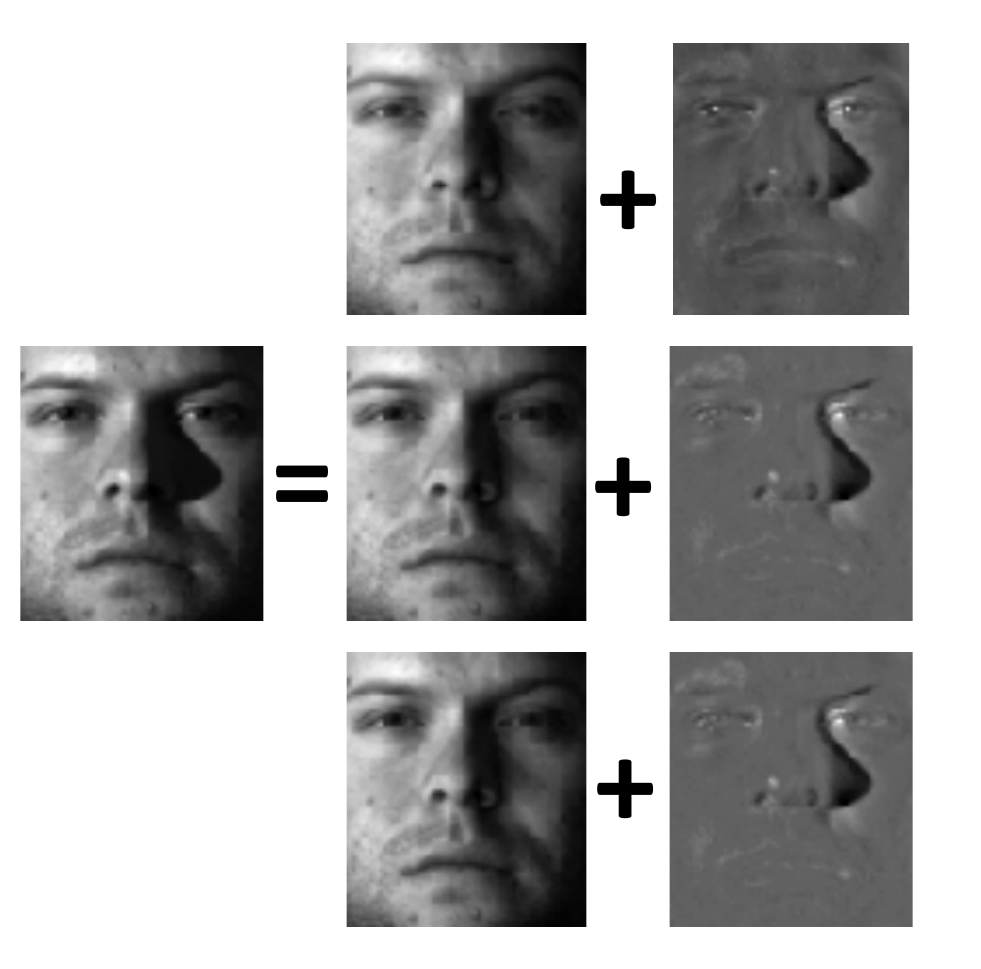}}
~~~~
    \subfloat[Subject 2]{
        \includegraphics[width=0.35\textwidth]{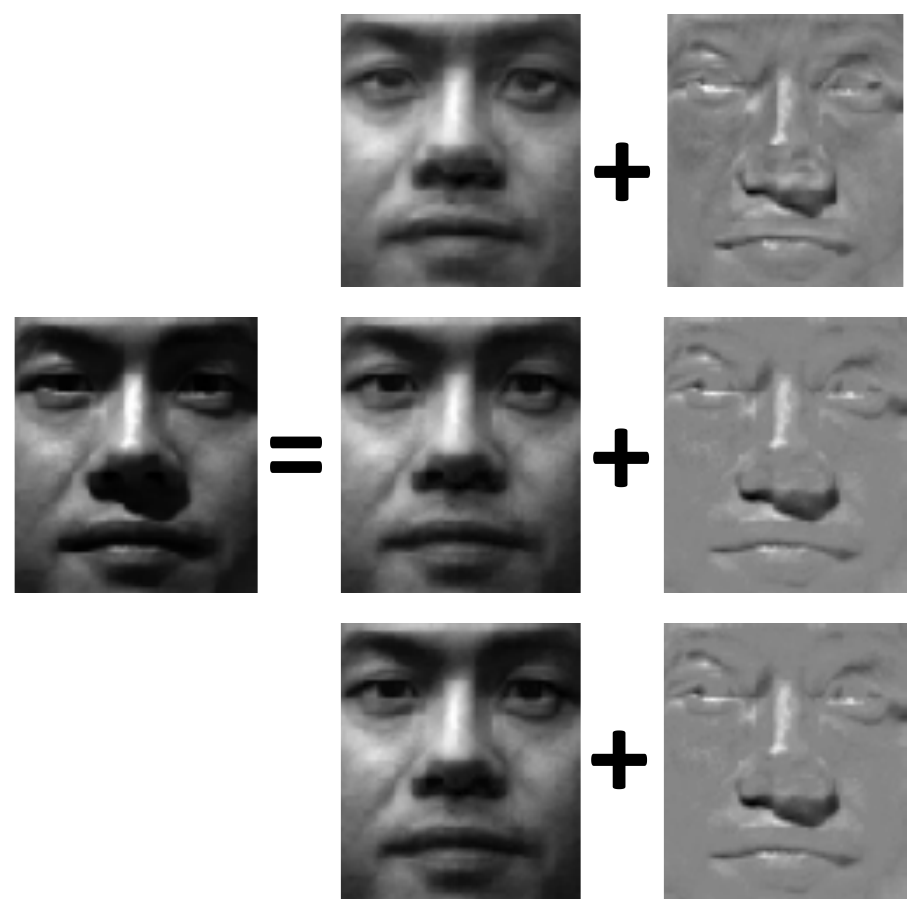}}\\
   \subfloat[Subject 3]{
		\includegraphics[width=0.35\textwidth]{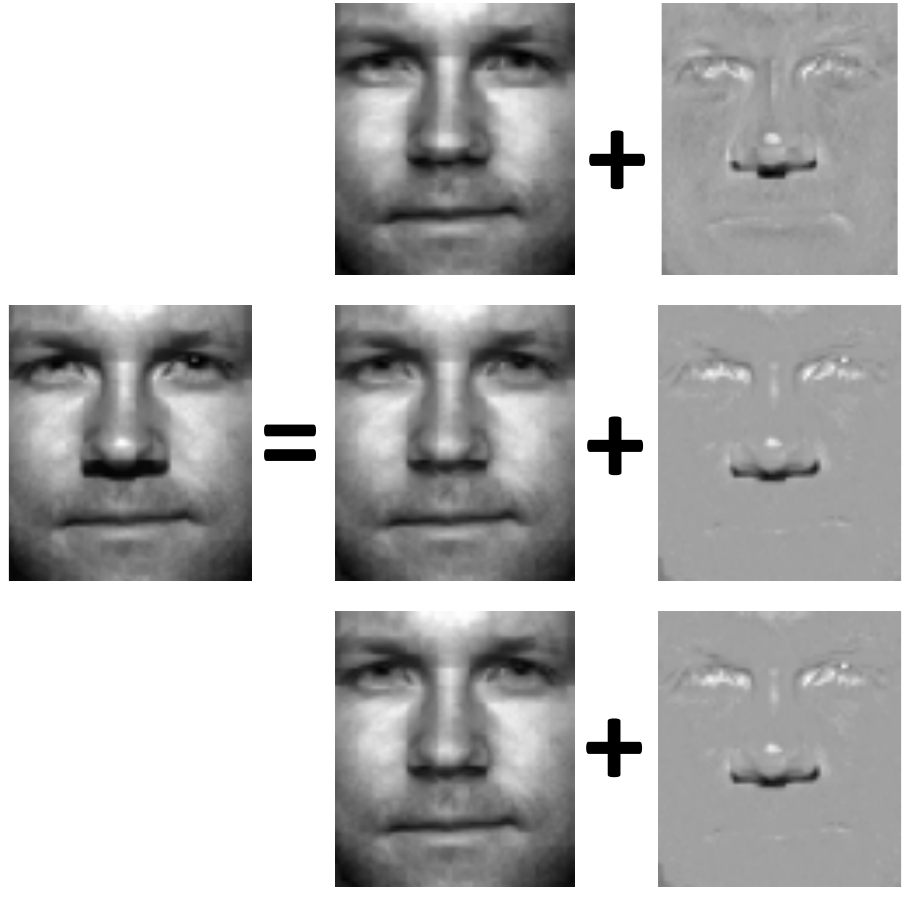}}
~~~~~~
       \subfloat[Subject 4]{
        \includegraphics[width=0.35\textwidth]{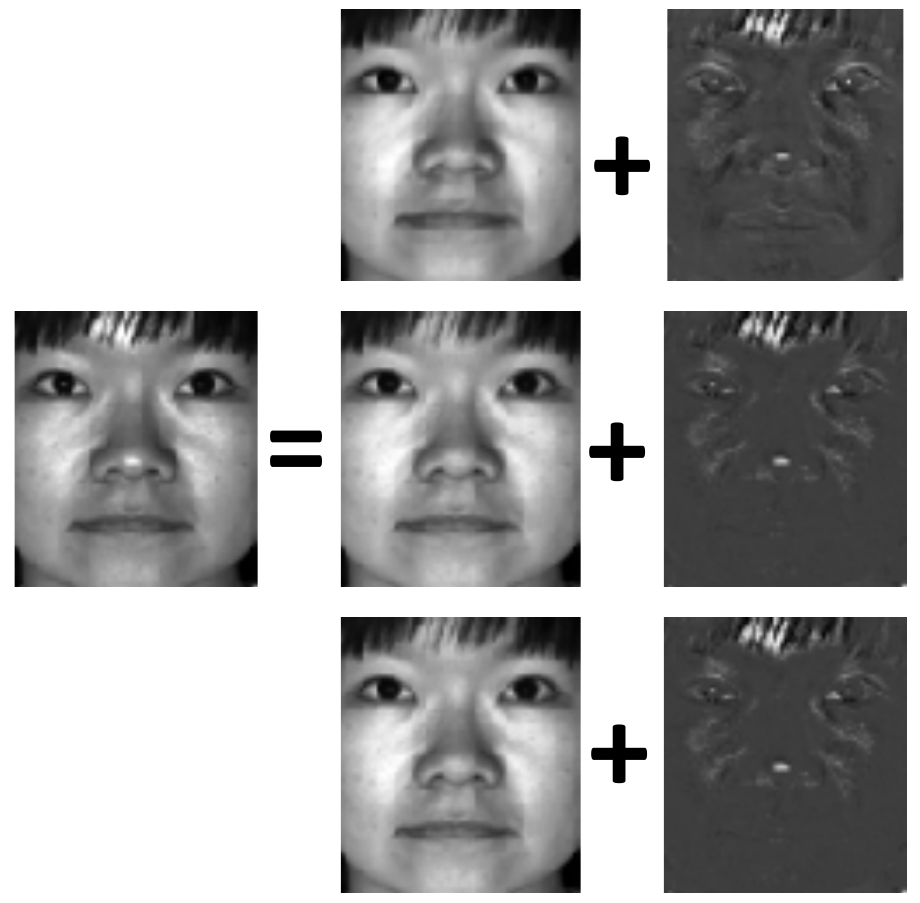}}
     %\vspace{-1em}
      \caption{Shadow removal results for facial images of Subject 1 (a) and Subject 2 (b); Light removed from facial images of Subject 3 (c) and Subject 4 (d). All the images are from the YaleB database. In each subfigure, the first row is the recovered low rank and sparse part by NSA. The second and third rows are the results of SPCP and our method.}	
      \label{fig_face}
\vspace{-1em}
\end{figure*}
%------------------------------------------------------------------------
%\subsection{Shadow/Light Removal from Face Images}
We further apply the principal component pursuit algorithms for removing light and shadow on facial images. Such processing is usually very important for face recognition. The reason that PCP can be applied for removing light/shadow on a face is that the captured facial image can be regarded as the sum of the low rank part (common face) and sparse errors (e.g., light/shadow). We test the competing algorithms on the YaleB data set, which consists of 38 subjects under different illumination. Each subject has 64 images with resolution $192 \times 168$. Thus the size of the observed matrix is $32256 \times 2432$. We apply NSA, SPCP and $p,q$-PCP on this data set, and plot four example images (1 per individual) in Figure \ref{fig_face}. It can be seen that the shadow and light on the faces are removed, and the recovered facial images are very clear. This verifies the effectiveness of our proposed algorithm. The three methods perform well in this situation.
%-------------------------------------------------------------------------
\section{Conclusions and Future Work}
\label{sec_5}
This study investigates the use of the Schatten-$p$ norm and the $\ell_q$-norm regularized non-convex principal component pursuit. We further develop an iteratively reweighted algorithm PIRA to solve the non-convex problem. In each iteration, PIRA solves two sub-problems which have closed form solutions. The obtained $L$ and $R$ are optimal for the linearized sub-problem. We demonstrate the convergence of the objective function based on the calculated low rank and sparse parts by experiments. %Theoretically analysis show that PIRA leads to the local minimum. At last,
We present extensive experiments on both synthetic and real-world data to demonstrate the attractive properties and effectiveness of our algorithm.
Interesting future work will be applying the joint Schatten-$p$ norm and $\ell_q$-norm to other low-rank and sparse problems in the area of video denoising.

\clearpage
\bibliographystyle{alpha}
\bibliography{pqPCP}
\end{document}